%% file: main.tex
\PassOptionsToPackage{table,dvipsnames}{xcolor}
\documentclass[]{selfevolagent}
\usepackage{microtype}
\usepackage{amsfonts}
\usepackage{amssymb}
\usepackage{xcolor}
\definecolor{lavender}{RGB}{162, 185, 250}
\definecolor{softlavender}{RGB}{216, 226, 251}
\definecolor{selfevolagent}{RGB}{117, 150, 248} 
\usepackage{graphicx}
\usepackage{booktabs}
\usepackage{wrapfig}
\usepackage{siunitx} 
\usepackage{float}
\usepackage{amsmath}
\usepackage{amsthm}
\usepackage{subcaption}
\usepackage{makecell}
\usepackage{algpseudocode}
\usepackage{algorithm}
\theoremstyle{plain}

\theoremstyle{definition}

\theoremstyle{remark}

\usepackage{enumitem}
\usepackage{pifont}
\usepackage[edges]{forest}
\usetikzlibrary{arrows.meta}
\usepackage{tikz}
\usepackage{graphicx}     
\usepackage{booktabs}     
\usepackage{enumitem}     
\usepackage[most]{tcolorbox} 
\usepackage{fontawesome5}
\usepackage{booktabs} 
\usepackage{url}
\usepackage{wrapfig}
\usepackage{longtable}
\usepackage{tabularx}
\usetikzlibrary{
  positioning,
  calc,
  shapes.symbols,
  shapes.geometric,
  shapes.misc
}
\usepackage[T1]{fontenc}
\definecolor{selfevolagent_dark}{HTML}{1B42E0}   
\definecolor{selfevolagent_light}{HTML}{5276F6}  
\definecolor{selfevolagent_lighter}{HTML}{EDF1FA}
\hypersetup{linkcolor=red}

\usepackage{booktabs}
\usepackage[table]{xcolor}
\usepackage[most]{tcolorbox}

\definecolor{rowlight}{RGB}{238, 240, 252}
\definecolor{rowdeep}{RGB}{216, 220, 248}

\newtcolorbox{takeawaybox}{
  colback=rowlight,
  colframe=blue!45!black,
  boxrule=0.5pt,
  arc=2pt,
  left=5pt,
  right=5pt,
  top=4pt,
  bottom=4pt,
  boxsep=1pt,
  before skip=6pt,
  after skip=6pt
}

\usepackage{listings}
\usepackage{xcolor}

\definecolor{codebg}{RGB}{245,245,245}
\definecolor{codecomment}{RGB}{0,128,0}
\definecolor{codekeyword}{RGB}{220,20,60}

\lstdefinestyle{pytorchstyle}{
    backgroundcolor=\color{codebg},   
    commentstyle=\color{codecomment},
    keywordstyle=\color{codekeyword},
    numberstyle=\tiny\color{gray},
    basicstyle=\ttfamily\footnotesize,
    breaklines=true,                 
    keepspaces=true,                 
    numbers=left,                    
    numbersep=5pt,                  
    showspaces=false,                
    showstringspaces=false,
    tabsize=4,
    language=Python
}

\usepackage{svg}
\usepackage{xltabular}
\usepackage{booktabs}
\usepackage{caption}
\usepackage{fontawesome5}
\usepackage{booktabs} 
\usepackage{url}
\usepackage{wrapfig}
\usepackage{longtable}
\usepackage{tabularx}
\usepackage{etoc}

\newcommand{\ghlink}[1]{\faIcon{github}\,\href{#1}{GitHub}}
\newcommand{\weblink}[1]{\faIcon{globe}\,\href{#1}{Website}}

\definecolor{rootcolor}{RGB}{27, 66, 224}
\definecolor{catcolor}{RGB}{82, 118, 246} 
\definecolor{subcatcolor}{RGB}{145, 172, 250} 
\definecolor{papercolor}{RGB}{237, 241, 250}

\title{\centering\textcolor{selfevolagent_dark}{MemOnce: A Scalable Memory Architecture for Large Language Models}}
\author[1]{Chang Nie}
\author[1]{Chaoyou Fu}
\author[2]{Songlin Liu}
\author[2]{Yiqing Hu}
\author[1]{Caifeng Shan}
\affiliation[1]{Nanjing University}
\affiliation[2]{University of Science and Technology of China}

\abstract{
\input{1_abstract}

}
\date{\today}
\metadata[\faEnvelope\ Main Contact]{changnie@smail.nju.edu.cn, bradyfu24@gmail.com}
\metadata[{\raisebox{-0.2ex}{\includegraphics[height=1em]{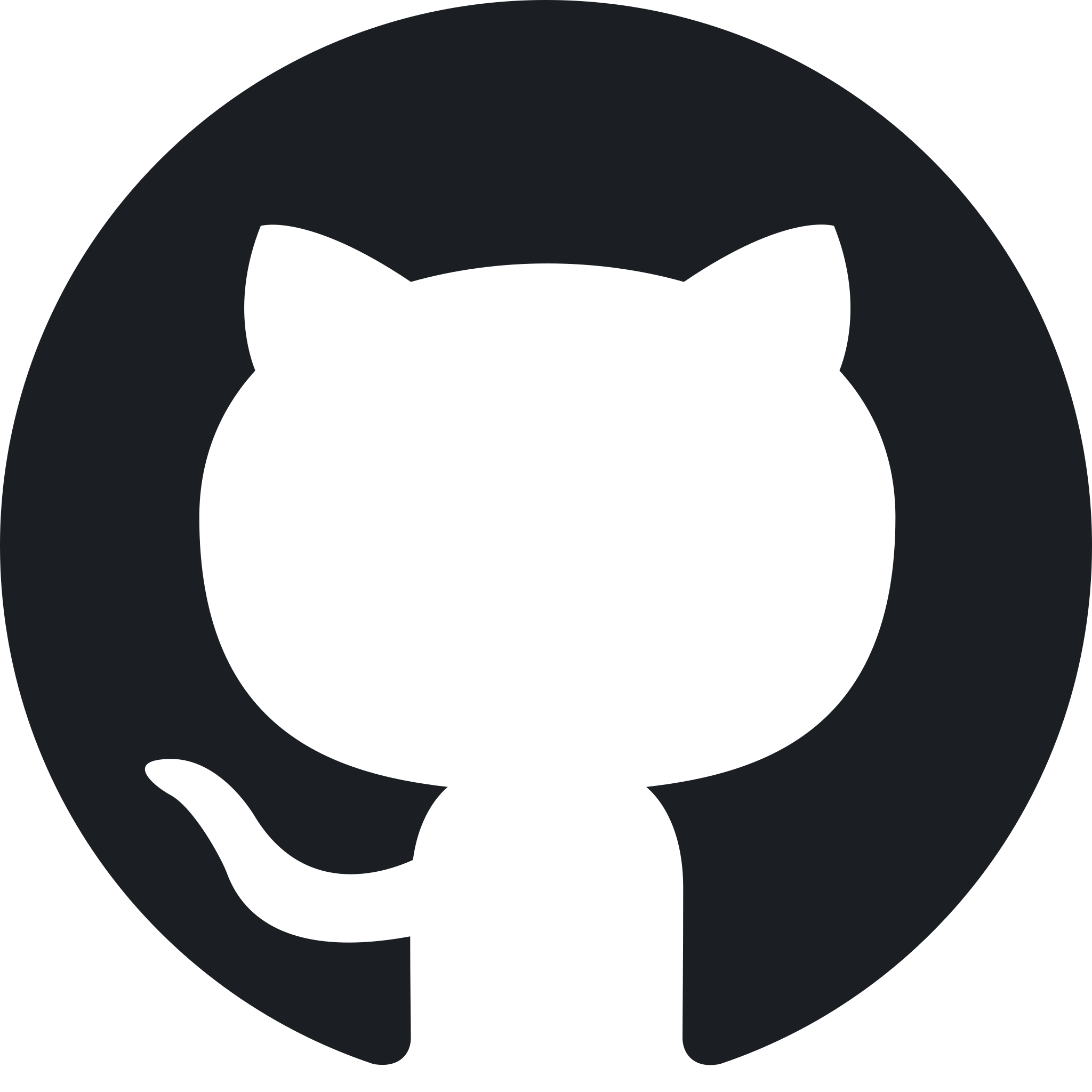}}\ Github}]{\url{https://clare-nie.github.io/MemOnce}}

\begin{document}

\vspace*{-2.5em}
\noindent
\begin{minipage}[c]{0.55\textwidth}
  \includegraphics[height=1cm]{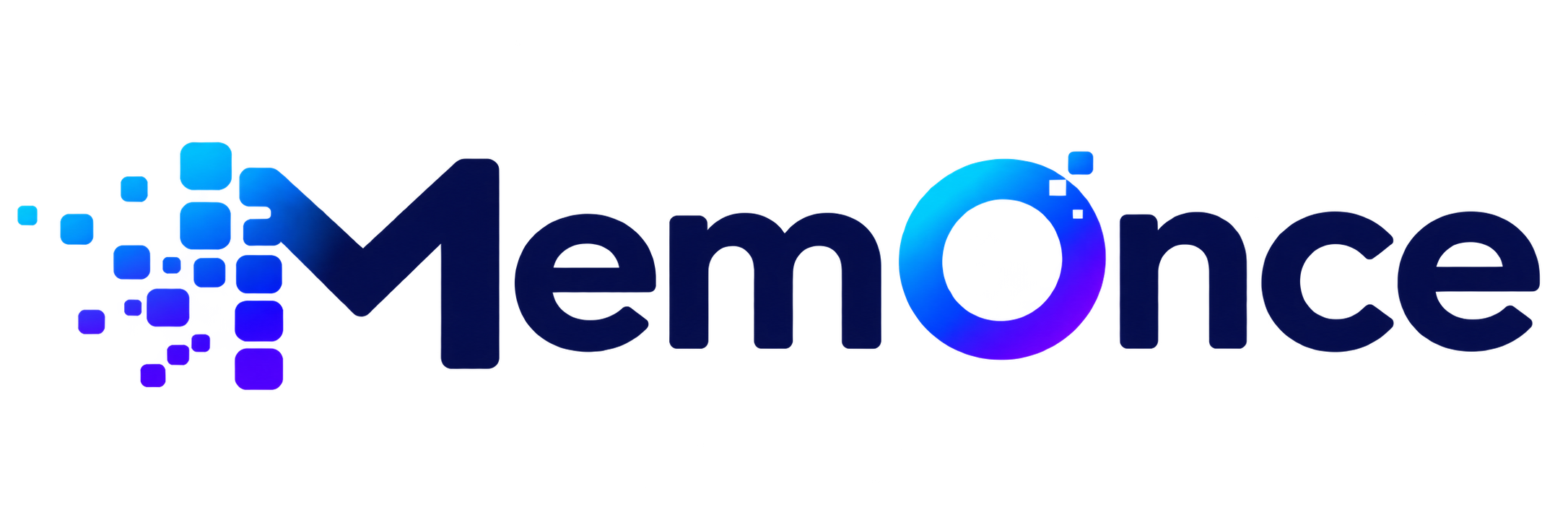}
\end{minipage}%
\hfill
\begin{minipage}[c]{0.4\textwidth}
  \raggedleft
\end{minipage}
\par
\vspace{0.5em}

\maketitle

\input{body}

\clearpage
\bibliographystyle{plainnat}
\bibliography{citation}
\clearpage
\appendix
\input{appendix}

\end{document}

%% file: 1_abstract.tex
\begin{abstract}

Deploying large language models (LLMs) in long-horizon and continual interactions requires maintaining reusable \textit{memory states} for future inference, posing challenges for existing architectures. 
Standard Transformers accumulate layer-wise, token-level KV caches, incurring linear GPU memory growth and quadratic computational complexity as the context expands. 
In contrast, linear-time architectures compress history into fixed-size states, but struggle to retain fine-grained details. 
We introduce \textbf{MemOnce}, a foundation architecture that learns \textit{scalable} memory states for language models. 
MemOnce is based on two core designs: 
(a) \textbf{Memory encoder-decoder}: it caches only a small subset of final-layer hidden states as compact memory representations and dynamically decodes them back into KV caches for attention layers;
(b) \textbf{Memory sparse attention}: for each continual input, MemOnce adaptively retrieves relevant memory representations, with the sparse indices computed only once across layers.
Coupling these two designs enables efficient on-the-fly memory loading and KV cache decoding, drastically reducing the GPU memory footprint and achieving linear-time complexity.
We conduct extensive experiments to show that MemOnce outperforms Transformers and other architectures across varying model sizes, context lengths, and training paradigms (pretraining and post-training adaptation).
In particular, adapting Qwen3.5-4B with MemOnce reduces memory state cache by \textbf{1313$\times$} and prefill latency by \textbf{6.9$\times$} at 1M context, and improves throughput by \textbf{3.5$\times$} and overall memory performance by \textbf{+5.77\%}.
Finally, MemOnce generalizes zero-shot to downstream representation tasks, achieving competitive results against specialized models.
\end{abstract}

%% file: body.tex
\input{2_introduction}

\input{3_preliminaries}

\input{4_mneme}

\input{5_experiments}

\input{6_conclusion}

%% file: 2_introduction.tex
\begin{figure*}[h]
\centering
\begin{minipage}[t]{.49\linewidth}
\centering
\vspace{0pt}
\includegraphics[width=\linewidth]{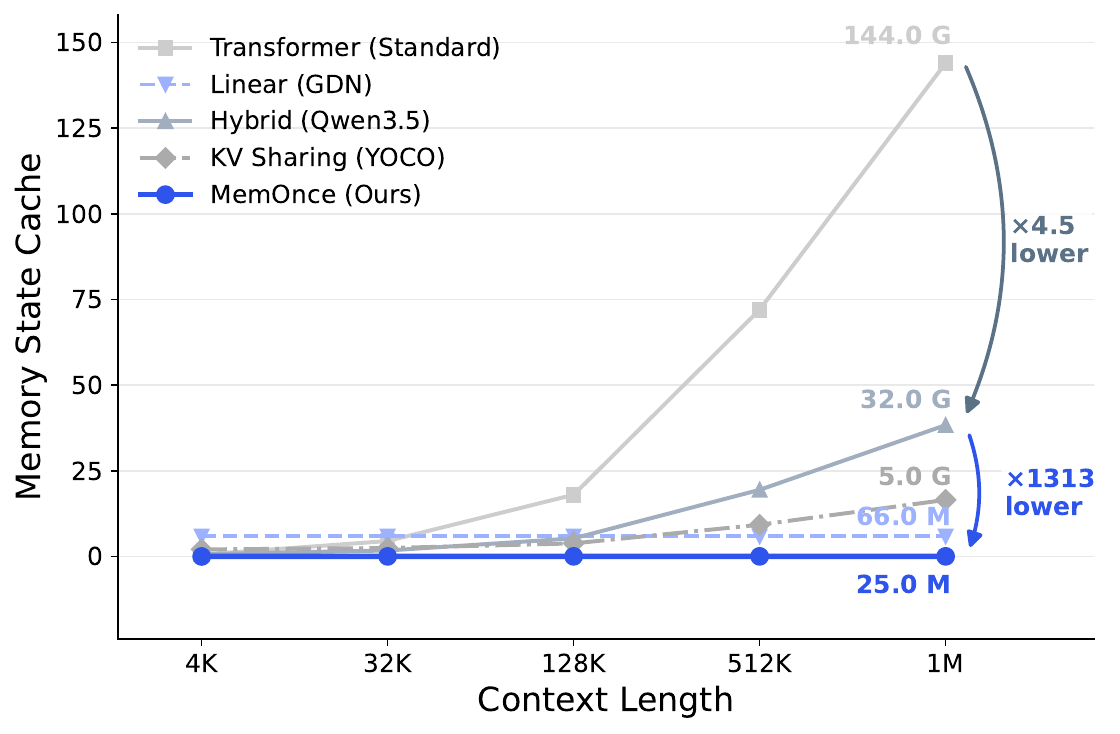}
\end{minipage}\hfill
\begin{minipage}[t]{.49\linewidth}
\centering
\vspace{0pt}
\includegraphics[width=\linewidth]{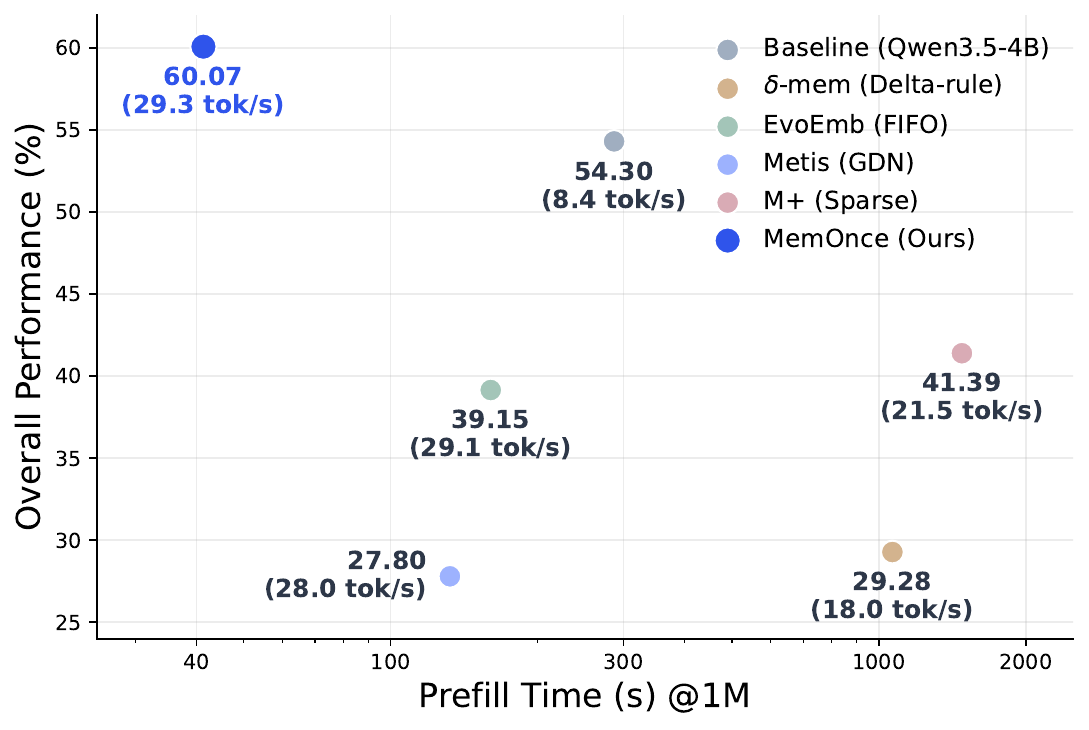}
\end{minipage}
\caption{\textbf{Comparison between MemOnce and advanced architectures.} 
\textbf{(Left)} Memory state cache vs.\ context length. MemOnce activates minimal memory states during continual inference, reducing the cache footprint by orders of magnitude compared to attention-based baselines. 
\textbf{(Right)} Overall performance vs.\ prefill latency and decoding throughput. Compared to existing native memory models, MemOnce significantly advances the Pareto frontier, marking the first native memory approach to surpass the full-context baseline. Performance is averaged across five memory benchmarks, while prefill time and throughput are measured at a 1M context length.\vspace{-2mm}}
\label{fig:overview}
\end{figure*}

\section{Introduction}
\label{sec:intro}

Memory serves as the foundation for experiential and continual intelligence~\citep{hassabis2017neuroscience,sumers2023cognitive,hu2025memory}.
As large language models (LLMs) transition from single-turn processing toward continual interactions and long-horizon agent tasks, maintaining reusable \textit{memory states} is essential to eliminate redundant context recomputation and ensure low-latency response generation~\citep{xi2025rise,chhikara2025mem0,zheng2026lifelong}. 
However, existing architectures face a fundamental trade-off between retention fidelity and inference scalability.
Transformers~\citep{vaswani2017attention} with global attention retain layer-wise, token-level key-value (KV) caches, incurring linear GPU memory growth and quadratic computational complexity as the context expands.
In contrast, recurrent and linear models (e.g., Mamba~\citep{gu2023mamba}, RWKV~\citep{peng2023rwkv}, and GDN~\citep{yang2025gated}) compress history into fixed-size states to enable constant-memory, linear-time inference; yet their bounded state capacity can induce information decay, hindering the preservation of fine-grained details over long horizons~\citep{jelassi2024repeat}.

Extensive efforts have sought to reconcile this trade-off, broadly falling into three categories:
(1) \textit{Architectural and attention optimizations}: hybrid models (e.g., Kimi K3~\citep{team2026kimi} and MiMo-V2.6~\citep{qiao2026mimov26}) interleave attention and linear layers, cross-layer sharing approaches~\citep{sun2024you,xu2026deepseek} reuse intermediate KV caches, and sparse attention mechanisms (e.g., DeepSeek-V4.1-Flash~\citep{xu2026deepseek} and Qwen3.8-Next~\citep{qiu2026design}) restrict attention to top-$k$ relevant tokens. 
However, during continual inference, they still need to maintain or access global caches that grow linearly with context length.
(2) \textit{Native memory models}: parametric memory designs (e.g., Titans~\citep{behrouz2025titans}, M+~\citep{wang2025m}, and Metis~\citep{zhang2026metis}) compress historical context into compact memory representations via specialized training or post-training adaptation. Yet, they often suffer from performance degradation, layer-wise retrieval overhead, or catastrophic forgetting.
(3) \textit{External memory systems}: agentic memory frameworks (e.g., Mem0~\citep{chhikara2025mem0}, A-MEM~\citep{xu2026mem}, and PersonaVLM~\citep{nie2026personavlm}) organize historical context into explicit databases that can be managed, updated, and queried. Nevertheless, constructing and maintaining these discrete stores is costly~\citep{nie2026lightomni} and difficult to optimize end-to-end~\citep{wang2025memalpha}.

Despite these efforts, a central question remains open: \textbf{\textit{How can we build a foundational memory architecture (or adapt an existing one) to simultaneously achieve accurate recall, linear-time inference, and bounded GPU memory cache overhead for continual inference?}}
In this paper, we propose \textbf{MemOnce}, an expressive yet scalable architecture that learns reusable memory states for language models. 
The core idea of MemOnce is to maintain a single compact global cache and retrieve relevant memory only once during continual inference to decode layer-wise KV caches, thereby enabling memory on-demand loading and decoding.
Specifically, we first develop a \textbf{memory encoder-decoder} that caches a small subset (e.g., $1/8$) of final-layer hidden states as compact memory representations, dynamically decoding them via lightweight MLPs into KV caches for attention layers.
This design achieves cache recurrence and an $Lr\times$ compression across both depth and time dimensions.
Second, we introduce \textbf{memory sparse attention}, which adaptively selects the top-$K$ task-relevant states for each input and shares the sparse routing indices across all attention layers and decoding steps.
By coupling these two designs, MemOnce achieves strong expressiveness, linear-time complexity, and a constant active memory state (see Figure~\ref{fig:overview}).

We conduct extensive experiments across diverse tasks, model sizes, and context scales to demonstrate the superiority of MemOnce. 
Specifically, we highlight three key results:
\textbf{(1) MemOnce as a foundation architecture}:
We train MemOnce and other foundation architectures from scratch across model sizes ranging from 150M to 1B parameters.
MemOnce consistently achieves lower or comparable perplexity on validation data and performs well on downstream benchmarks compared to evaluated baselines (e.g., Transformer, GDN, and YOCO~\citep{vaswani2017attention,yang2025gated,sun2024you}).
\textbf{(2) MemOnce as a native memory model}:
We adapt Qwen3.5 via post-training and evaluate it on five memory benchmarks spanning 4K to 1M tokens. As shown in Figure~\ref{fig:overview}, MemOnce is the first native memory model to surpass the full-context baseline (+5.77\% at the 4B scale)~\citep{qwen35blog}, and outperforms prior memory methods (by +18.68\% over M+~\citep{wang2025m} and +32.27\% over Metis~\citep{zhang2026metis}).
At 1M context, it reduces memory state cache by 1313$\times$ and prefill latency by 6.9$\times$, and improves throughput by 3.5$\times$.
Moreover, MemOnce avoids catastrophic forgetting on general benchmarks by keeping its parameters frozen during inference and inherently ignoring irrelevant decoded caches.
\textbf{(3) MemOnce as a representation model}:
We show that MemOnce's learned memory representations transfer naturally to downstream retrieval and reranking tasks, outperforming specialized models without task-specific fine-tuning. 
These findings underscore the potential of MemOnce as a next-generation general-purpose architecture.

In summary, our main contributions are as follows:
\begin{itemize}
    \item We propose \textbf{MemOnce}, a foundation architecture that achieves strong expressiveness, linear-time complexity, and constant active memory for continual inference.
    \item We couple two designs, memory encoder-decoder and memory sparse attention, to cache and retrieve memory representations only once for on-demand memory loading and KV cache decoding.
    \item We establish through comprehensive experiments that MemOnce advances the performance--complexity Pareto frontier, marking the first native memory model to surpass the full-context baseline and generalizing zero-shot to downstream representation tasks.
\end{itemize}

%% file: 3_preliminaries.tex
\section{Preliminaries}
\label{sec:preliminaries}

\paragraph{Problem Setup.}
Consider an $L$-layer autoregressive language model $\pi_\theta$ with hidden dimension $D$ processing an input stream $x_1, x_2, \dots, x_t$ (e.g., token/chunk sequences in language modeling or interaction turns in continual tasks~\citep{zheng2026lifelong}). At step $t$, the model receives the current input $x_t$ alongside accumulated historical context $x_{1:t-1}$ of total token length $N$ ($N \geq t-1$), and generates the target output $y_t$. 
In each layer $l \in \{1, \dots, L\}$, let $H^l = [h_1^l, h_2^l, \dots, h_{t-1}^l]^\top \in \mathbb{R}^{N \times D}$ denote the accumulated hidden states produced for $x_{1:t-1}$ (where each $h_i^l$ corresponds to a token or segment, totaling $N$ tokens), with $H^0$ representing the token embeddings and $H^L$ denoting the output of the final attention layer.

\paragraph{Unified Perspective: Memory State Learning.}
We define \textit{Memory State Learning} as the problem of constructing reusable, active \textit{memory states} $M_t$ for the current input $x_t$ during continual inference, so as to avoid recomputing the raw history $x_{1:t-1}$ and ensure low-latency response generation.
Formally, we reformulate the context-conditioned generation of existing methods into generation conditioned on the active memory state $M_t$:
\begin{equation}
    \begin{aligned}
        \pi_\theta(y_t \mid x_t, x_{1:t-1}) &= \pi_\theta(y_t \mid x_t, M_t), \\
        \text{where} \quad M_t &\doteq f_\theta(x_t, x_{1:t-1}).
    \end{aligned}
    \label{eq:memory_state_learning}
\end{equation}
Here, $f_\theta(x_t, x_{1:t-1})$ denotes the memory state mapping that distills the historical context $x_{1:t-1}$ and activates relevant states $M_t$ with respect to the current input $x_t$.
Under Eq.~\eqref{eq:memory_state_learning}, existing methods correspond to distinct instantiations of $f_\theta$:
\begin{itemize}
    \item \textbf{Multi-Layer Token KV Caches (Transformers):} 
    Standard Transformers~\citep{vaswani2017attention} define a query-agnostic $f_\theta$ by activating all token-level key-value caches across all $L$ layers:
    \begin{equation}
        \begin{aligned}
            &M_t^{\text{KV}} = \Big\{\big(K^l, V^l\big)\Big\}_{l=1}^{L} \in \mathbb{R}^{L \times N \times 2D}, \\
            &K^l = H^{l-1}W_K^l, \quad V^l = H^{l-1}W_V^l.
        \end{aligned}
    \end{equation}
    While retaining fine-grained details, $M_t^{\text{KV}}$ grows linearly with both depth $L$ and context length $N$, incurring an $\mathcal{O}(LND)$ memory footprint. 
    Even for sparse attention~\citep{xu2026deepseek,qiu2026design} or retrieval-based memory models~\citep{wang2025m} that condition $f_\theta$ on $x_t$ to select top-$k$ context per layer, they still maintain multi-layer caches and compute routing indices across layers.

    \item \textbf{Online Associative States (Linear Models):} 
    Linear and recurrent models~\citep{gu2023mamba,yang2025gated,kimi_team_2025_kimi_linear} implement $f_\theta$ by compressing history into a fixed-size state $S_N^l$ at each layer for all subsequent inputs. Specifically, representative associative linear models (e.g., GDN~\citep{yang2025gated}) maintain a state matrix $S_i^l \in \mathbb{R}^{d \times d}$ updated via the online delta rule to encode key-value associations:
    \begin{equation}
        \begin{aligned}
            &M_t^{\text{Linear}} = \big\{S_N^l\big\}_{l=1}^{L} \in \mathbb{R}^{L \times d \times d}, \\
            & S_i^l = \alpha_i^l S_{i-1}^l + \beta_i^l \big(v_i^l - \alpha_i^l S_{i-1}^l k_i^l\big) {k_i^l}^\top,
        \end{aligned}
        \label{eq:linear_state}
    \end{equation}
    where $\alpha_i^l$ and $\beta_i^l$ denote the erase and write gates. Although $M_t^{\text{Linear}}$ achieves $\mathcal{O}(Ld^2)$ constant memory complexity (independent of $N$) and linear-time inference, it suffers from severe capacity bottlenecks~\citep{jelassi2024repeat}.

    \item \textbf{Cross-Layer Shared States (YOCO):} 
    Architectures with cross-layer KV sharing, such as YOCO~\citep{sun2024you} and Causal Encoder-Decoder (CED)~\citep{xu2026deepseek}, cache hidden states at an intermediate layer (e.g., $L/2$ in YOCO) to decode KV caches for subsequent layers:
    \begin{equation}
        M_t^{\text{YOCO}} = H^{L/2} \in \mathbb{R}^{N \times D}.
    \end{equation}
    This design eliminates the depth factor $L$, but $M_t^{\text{YOCO}}$ still grows linearly with context length.
\end{itemize}
For the above architectures, $f_\theta$ is implicitly learned during the optimization of $\pi_\theta$ and executed at test time.
Moreover, for native memory models~\citep{wang2025m,zhang2026metis}, $f_\theta$ reconstructed via post-training adaptation can also be subsumed under the aforementioned categories.
However, constructing an effective and scalable $M_t$ remains an open challenge (see Appendix~\ref{app:theory} for further theoretical analysis).

%% file: 4_mneme.tex
\begin{figure*}[!t]
    \centering
    \includegraphics[width=0.7\textwidth]{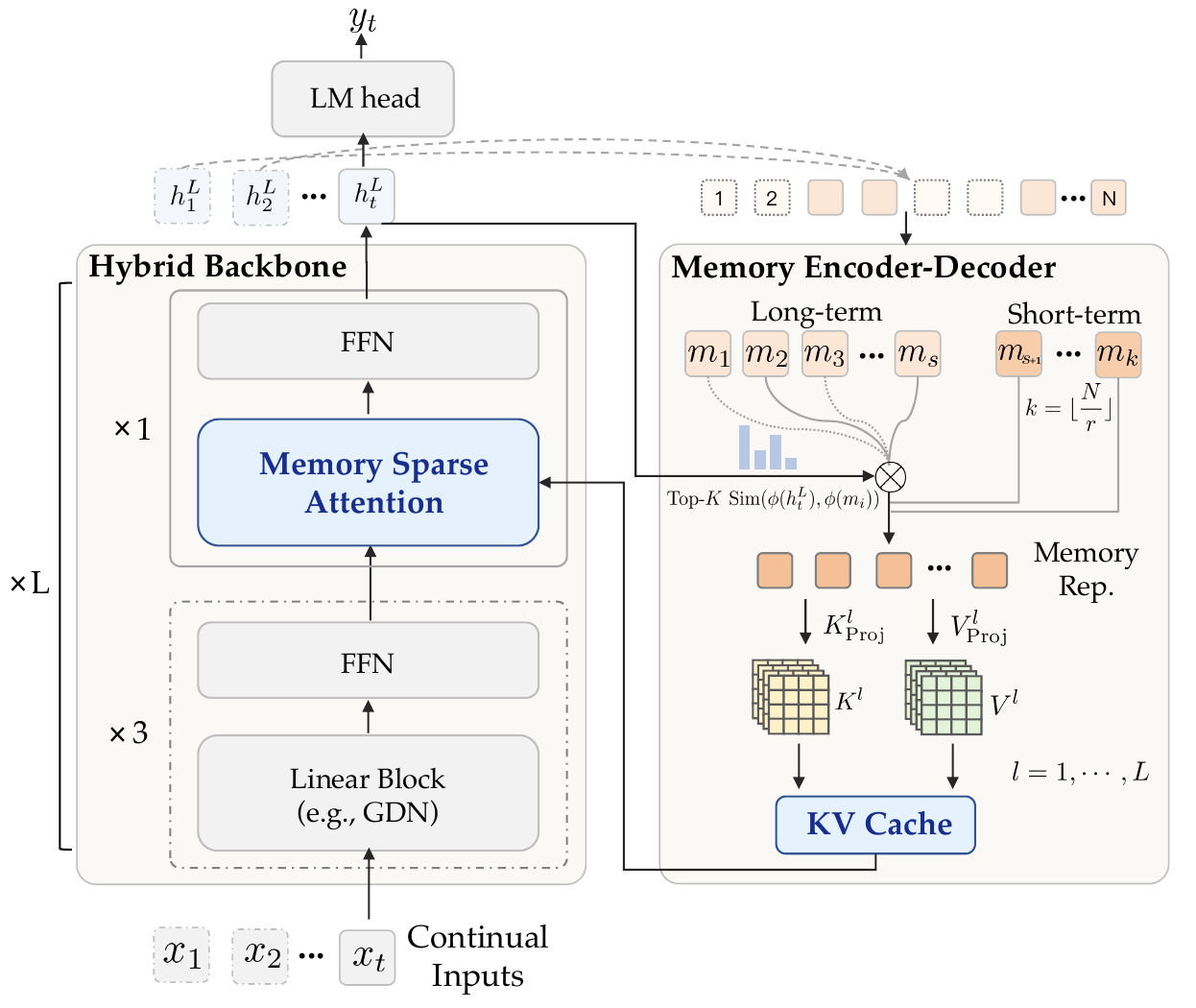}
\caption{\textbf{Overview of the MemOnce architecture.} 
\textbf{(Left)} The \textit{Hybrid Backbone} stacks linear blocks with memory sparse attention across $L$ stages. 
\textbf{(Right)} The \textit{Memory Encoder-Decoder} caches a trailing subset ($1/r$) of final-layer states into a persistent memory bank, partitioned into a local short-term window and a sparse long-term pool. 
For each incoming input, relevant memory states are retrieved only once and dynamically decoded into KV caches for attention layers. 
This enables a constant active memory state for continual inference, alongside an $Lr\times$ cache compression across both depth and time dimensions.}
    \label{fig:architecture}
\end{figure*}

\section{MemOnce}
\label{sec:method}

We introduce \textbf{MemOnce}, a foundation architecture that learns scalable memory states $M_t$ in Eq.~\eqref{eq:memory_state_learning} for efficient continual inference. 
Formally, MemOnce constructs and utilizes the active memory state $M_t$ through three functions:
\begin{equation}
    G_t = g_\theta(x_{1:t-1}), \ \  M_t = i_\theta(x_t, G_t), \ \  C_t = u_\theta(M_t),
    \label{eq:three_stages}
\end{equation}
where $g_\theta$, $i_\theta$, and $u_\theta$ denote the memory writing, indexing, and reading operations, respectively.
We instantiate Eq.~\eqref{eq:three_stages} through two core designs: 
(1) the \textbf{memory encoder-decoder} ($g_\theta$ and $u_\theta$), which caches partial final-layer states into a persistent memory bank $G_t$ and decodes the active memory state $M_t$ into multi-layer KV caches $C_t$ on demand; and 
(2) \textbf{memory sparse attention}, which retrieves the relevant memory state $M_t$ via $i_\theta$ only once and restricts historical attention exclusively to the decoded KV caches $C_t$.

MemOnce is applicable to various attention-based models.
In practice, as shown in Figure~\ref{fig:architecture}, MemOnce adopts an $L$-stage hybrid backbone by default, where each stage interleaves three linear blocks (e.g., GDN~\citep{yang2025gated}) with one \textit{Memory Sparse Attention} block.

\subsection{Memory Encoder-Decoder}
\label{subsec:encoder_decoder}

To eliminate the $\mathcal{O}(LND)$ storage bottleneck of multi-layer token caches, the memory encoder-decoder compresses historical context across both layer depth $L$ and sequence length $N$, and reconstructs layer-specific KV caches on demand.

\paragraph{Cache Memory Only Once ($g_\theta$).}
Inspired by prior cross-layer caching designs~\citep{sun2024you,xu2026deepseek}, MemOnce maintains a single global cache generated from final-layer hidden states, encoding historical context into compact memory representations at test time.
Specifically, given historical inputs $x_{1:t-1}$ of total length $N$, each segment $x_i$ of length $n_i$ ($\sum_{i=1}^{t-1} n_i = N$) is processed to produce final-layer hidden states $h_i^L \in \mathbb{R}^{n_i \times D}$.
Within each $x_i$, the interleaved linear blocks and causal attention continuously aggregate sequential context toward later positions, allowing the trailing token states in $h_i^L$ to encapsulate the semantics of the entire segment. 
Leveraging this property, the memory encoder $g_\theta$ retains only the last $1/r$ fraction of token states (denoted as $h_{i, \,-n_i/r:}^L$) from each $h_i^L$ with compression ratio $r$ (e.g., $r=8$), writing a total of $k = \lfloor \frac{N}{r} \rfloor$ compact memory representations into the global cache:
\begin{equation}
    \begin{aligned}
        G_t &= \operatorname{Concat}\Big(\big\{ h_{i, \,-n_i/r:}^L \big\}_{i=1}^{t-1}\Big) \\
        &= [m_1, m_2, \dots, m_k]^\top \in \mathbb{R}^{k \times D}.
    \end{aligned}
    \label{eq:memory_encoder}
\end{equation}
This step reduces the persistent memory by $Lr\times$, eliminating the layer multiplier $L$ and compressing the sequence dimension by $r$.
In practice, multiple segments are processed in parallel to accelerate prefilling.

\begin{figure*}[!th]
\centering
\includegraphics[width=.49\linewidth]{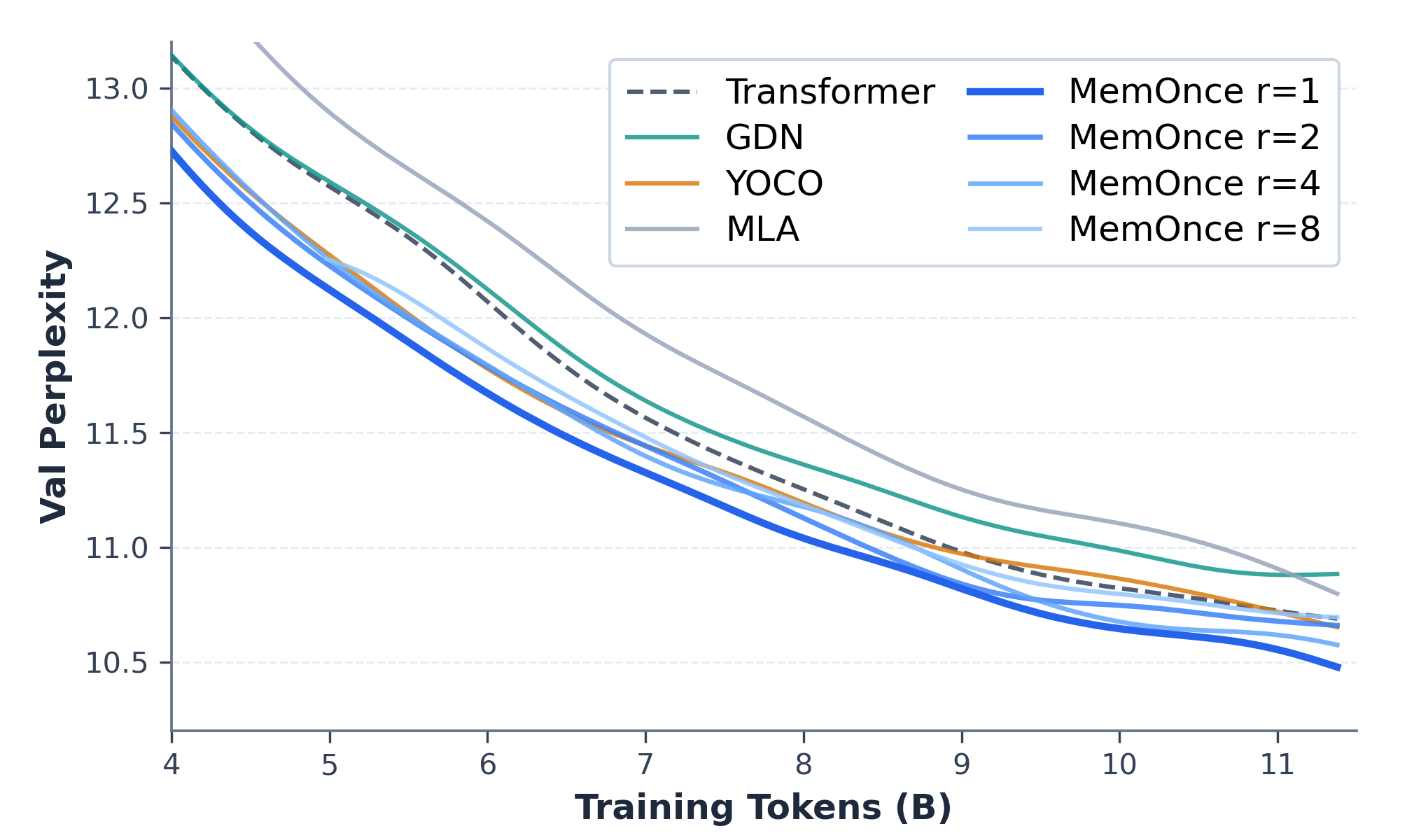}
\includegraphics[width=.49\linewidth]{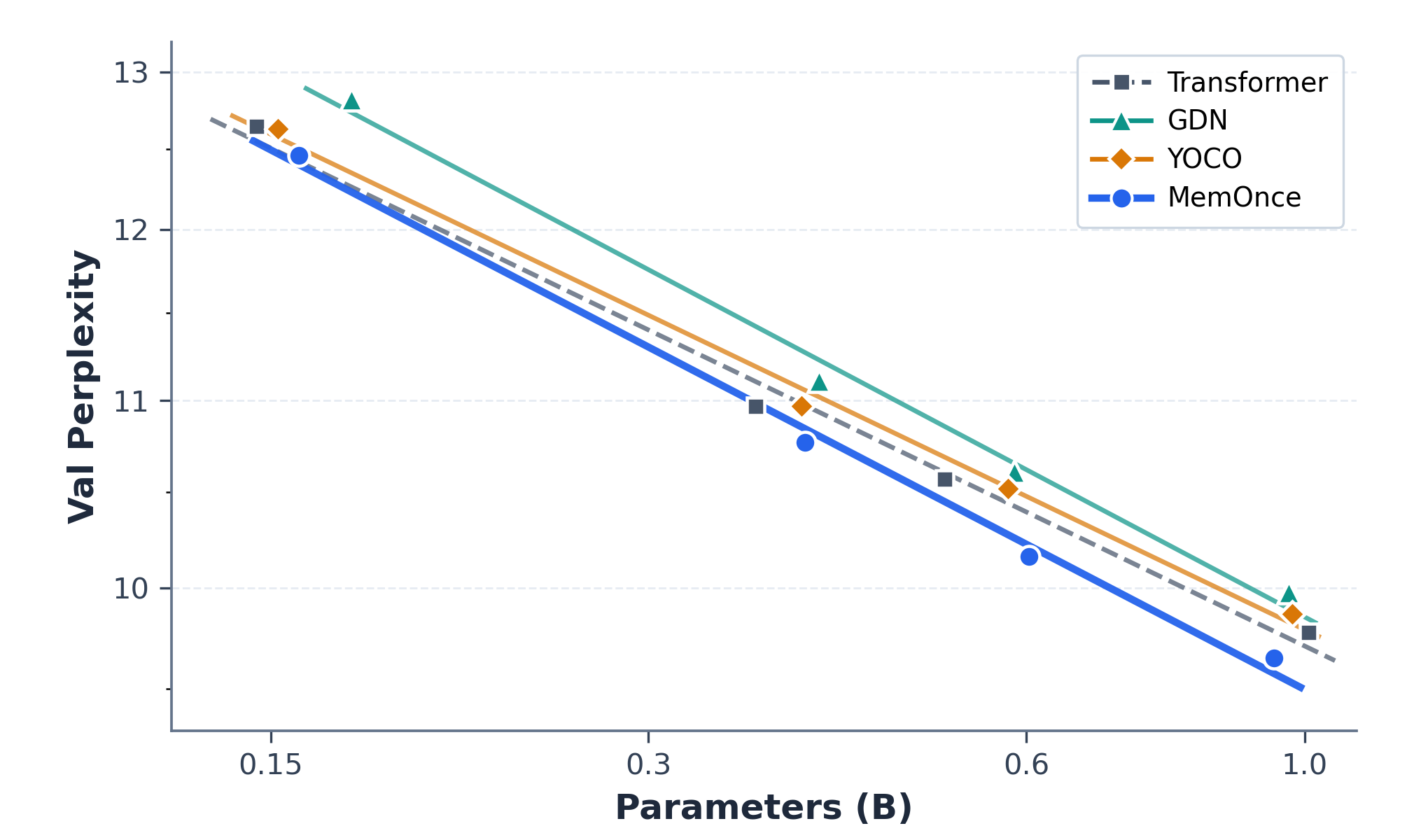}
\caption{\textbf{Training and scaling curves of MemOnce as a foundation architecture.}
\textbf{(Left)} Validation perplexity vs.\ training tokens under different compression ratios. Even at $r=8$, MemOnce remains competitive with the Transformer and strong baselines.
\textbf{(Right)} Validation perplexity vs.\ model scale (0.15B to 1.0B), where MemOnce delivers the best scaling performance.}
\label{fig:encoder_decoder}
\end{figure*}

\paragraph{Decode KV Caches on Demand ($u_\theta$).}
Directly sharing identical KV states across layers restricts the representational capacity of attention blocks at different depths. 
Instead, given the retrieved active memory state $M_t \in \mathbb{R}^{|M_t| \times D}$, the memory decoder $u_\theta$ applies lightweight layer-specific MLP projections $K_{\text{Proj}}^l$ and $V_{\text{Proj}}^l$ to decode $M_t$ into KV caches for all $L$ attention layers on demand:
\begin{equation}
    K^l = K_{\text{Proj}}^l(M_t), \ \  V^l = V_{\text{Proj}}^l(M_t), \ \  l \in \{1, \dots, L\}.
    \label{eq:memory_decoder}
\end{equation}
By projecting final-layer semantic representations back to all attention blocks through layer-specific decoders, MemOnce establishes looped memory generation and utilization.
As shown in Figure~\ref{fig:encoder_decoder}, across different compression ratios $r$ (even at $r=8$) and model scales, MemOnce achieves comparable or lower validation perplexity compared to the Transformer and strong baselines~\citep{vaswani2017attention,sun2024you,yang2025gated}.

\subsection{Memory Sparse Attention}
\label{subsec:sparse_attention}

While the memory encoder-decoder substantially compresses the global cache by $Lr\times$, decoding the entire memory bank $G_t$ across ultra-long horizons would still incur linear growth in cache memory and attention compute.
We introduce memory sparse attention to break fixed capacity limits while maintaining constant active states and linear-time complexity.
Its core idea is to retrieve relevant memory representations only once for each continual input and reuse them across layers with negligible loading overhead.

\paragraph{Retrieve Memory Only Once ($i_\theta$).}
As illustrated in Figure~\ref{fig:architecture}, MemOnce partitions the persistent memory bank $G_t = [m_1, \dots, m_k]^\top$ into a long-term memory pool $G_t^{\text{long}} = \{m_1, \dots, m_s\}$ and a short-term memory window $G_t^{\text{short}} = \{m_{s+1}, \dots, m_k\}$. 
The short-term window retains the most recent $k - s$ memory representations to preserve local continuity, whereas the long-term pool stores distant historical states that are accessed sparsely. 
Given the final-layer representation $h_t^L$ of the continual input $x_t$, the indexing function $i_\theta$ selects the top-$K$ task-relevant memory states from $G_t^{\text{long}}$ only once:
\begin{equation}
    M_t^{\text{long}} = \operatorname{Top-K}_{m_i \in G_t^{\text{long}}}\Big(\operatorname{Sim}\big(\phi(h_t^L), \phi(m_i)\big)\Big),
    \label{eq:sparse_retrieval}
\end{equation}
where $\phi(\cdot)$ extracts the last-token representation and projects it via an MLP layer followed by $\ell_2$-normalization, $\operatorname{Sim}(\cdot, \cdot)$ computes cosine similarity, and $\operatorname{Top-K}(\cdot)$ selects the $K$ most relevant memory representations to bound the active memory budget.
The retrieved long-term representations are concatenated with the short-term window to form the active memory state $M_t$:
\begin{equation}
    M_t = i_\theta(x_t, G_t) = \big[ M_t^{\text{long}} \,\|\, G_t^{\text{short}} \big] \in \mathbb{R}^{|M_t| \times D}.
    \label{eq:active_memory_subset}
\end{equation}

\paragraph{Sparse Attention and On-Demand Loading.}
Once $M_t$ is retrieved and decoded into layer-specific KV caches via Eq.~\eqref{eq:memory_decoder}, each attention block $l \in \{1, \dots, L\}$ prepends the decoded $(K^l, V^l)$ to the current input's local key-value states $(K_t^l, V_t^l)$ for attention computation:
\begin{equation}
    O_t^l = \operatorname{Attn}\Big(Q_t^l, \, [K^l \| K_t^l], \, [V^l \| V_t^l]\Big),
    \label{eq:memory_attn}
\end{equation}
where $Q_t^l, K_t^l, V_t^l$ denote the query, key, and value states of $x_t$ at stage $l$, and $[\cdot \| \cdot]$ denotes sequence concatenation.
Unlike layer-wise memory retrievers~\citep{wang2025m} and token-wise sparse attention mechanisms~\citep{yuan2025native,xu2026deepseek} that repeatedly compute routing indices, MemOnce performs Eq.~\eqref{eq:sparse_retrieval} only once per query and reuses $M_t$ across all $L$ attention blocks and subsequent decoding steps.
Because the active subset size $|M_t| = C \ll k$ is bounded by a constant budget, MemOnce restricts the active memory state cache to $\mathcal{O}(CD)$ (i.e., $\mathcal{O}(1)$ with respect to context length $N$ and depth $L$) while maintaining an expandable persistent bank $G_t$ that can be offloaded to CPU memory or SSD storage.\footnote{In continual inference, only the active memory state $M_t \in \mathbb{R}^{C \times D}$ is maintained in GPU cache as historical context cache.}
In practice, memory retrieval and loading in MemOnce take only $64.2\text{ ms}$ end-to-end, which is over $12\times$ faster than layer-wise retrieval in M+ and over $17\times$ faster than RAG pipelines~\citep{ma2023query} (see Table~\ref{tab:ablation_efficiency} in Appendix~\ref{sec:ablation}).

\subsection{Analysis and Optimization}
\label{subsec:analysis_optimization}

\paragraph{Advantage Analysis.}
Compared with existing foundation architectures and native memory models, MemOnce offers three key advantages:
\textbf{(1) Strong Performance:} By maintaining an expandable memory bank $G_t$ and retrieving final-layer representations back into multi-layer KV caches, MemOnce overcomes fixed-capacity bottlenecks and achieves accurate recall in long-context scenarios.
\textbf{(2) Bounded Active Memory State:} MemOnce reduces the persistent cache footprint by a factor of $Lr$ and achieves constant active memory state during continual inference.
\textbf{(3) High Inference Efficiency:} As illustrated in Figure~\ref{fig:overview} (right), MemOnce operates with linear-time complexity and achieves superior inference efficiency over existing memory architectures.


\begin{wraptable}{r}{0.5\textwidth}
\centering
\captionsetup{font=small}
\caption{\textbf{Memory state cache and prefilling time complexities.} $L$: layer depth, $N$: context length, $D$: hidden dimension, and $C$: an independent constant ($C \ll N$).}
\label{tab:complexity}
\setlength{\tabcolsep}{2pt}
\resizebox{0.48\textwidth}{!}{%
\begin{tabular}{lcc}
\toprule
\textbf{Model} & \textbf{State Cache} & \textbf{Prefilling Time} \\
\midrule
Transformer~\citep{vaswani2017attention} & $\mathcal{O}(LND)$ & $\mathcal{O}(LN^2D)$ \\
GDN~\citep{yang2025gated} & $\mathcal{O}(CLD)$ & $\mathcal{O}(CLND)$ \\
SWA~\citep{beltagy2020longformer} & $\mathcal{O}(CLD)$ & $\mathcal{O}(CLND)$ \\
YOCO~\citep{sun2024you} & $\mathcal{O}(ND)$ & $\mathcal{O}(LN^2D)$ \\
M+~\citep{wang2025m} & $\mathcal{O}(CLD)$ & $\mathcal{O}(CLND)$ \\
\midrule
\textbf{MemOnce} & $\mathcal{O}(CD)$ & $\mathcal{O}(CLND)$ \\
\bottomrule
\end{tabular}%
}
\end{wraptable}

\paragraph{Complexity Comparison.}
Table~\ref{tab:complexity} reports the memory state cache and prefilling time complexities of representative architectures.
Standard Transformers and YOCO~\citep{sun2024you} retain full-sequence caches ($\mathcal{O}(LND)$ and $\mathcal{O}(ND)$) and require quadratic prefilling time of $\mathcal{O}(LN^2D)$. 
Linear GDN~\citep{yang2025gated}, SWA~\citep{beltagy2020longformer}, and M+~\citep{wang2025m} reduce prefilling time to $\mathcal{O}(CLND)$ but still maintain $L$-layer states of $\mathcal{O}(CLD)$.
Moreover, fixed states limit memory capacity and degrade long-range recall, whereas M+ incurs retrieval and loading overhead.
In contrast, MemOnce achieves linear prefilling time and strictly bounds the memory state cache to $\mathcal{O}(CD)$.

\paragraph{Optimization.}
MemOnce supports end-to-end optimization under the standard next-token prediction paradigm for both pretraining and post-training adaptation.
During pretraining, the backbone and memory encoder-decoder are jointly optimized, where historical segments are encoded in parallel into $G_t$ and gradients flow directly through the active memory state $M_t$ and the decoded KV caches.
During post-training adaptation, we freeze the backbone and optimize only the lightweight projections in $i_\theta$ and $u_\theta$, alongside low-rank adapters (LoRA) activated exclusively during memory encoding ($g_\theta$).
Because historical context is internalized as decoded KV caches and the backbone remains frozen during response generation, MemOnce equips pretrained LLMs with native long-range memory capabilities without compromising their general task performance. More details are provided in Appendix~\ref{app:memonce_analysis}.

%% file: 5_experiments.tex
\section{Experiments}
\label{sec:experiments}

\subsection{Experimental Setup}
\label{subsec:exp_setup}

\paragraph{Evaluation and Benchmarks.}
We conduct extensive experiments to evaluate MemOnce across diverse tasks, model scales, and context lengths.
Our evaluation is organized into four primary tracks:
\textbf{(1) Foundational Capability:} We evaluate language modeling performance using perplexity (PPL) on WikiText~\citep{merity2017pointer} and LAMBADA~\citep{paperno2016lambada}. We further report zero-shot accuracy across eight downstream tasks using the LM-Evaluation-Harness~\citep{gao2021framework} to assess general knowledge and reasoning capabilities.
\textbf{(2) Long-Context Memory Capability:} We evaluate continual and long-horizon memory recall across context lengths spanning 4K to 1M tokens on standard memory benchmarks, including MemBench~\citep{tan2025membench}, LoCoMo~\citep{maharana2024evaluating}, LongMemEval~\citep{wu2025longmemeval}, and PersonaMem~\citep{jiang2025know}.
\textbf{(3) General Capability Retention:} To measure general capability retention under active memory load and verify robustness against catastrophic forgetting, we evaluate models on MMLU~\citep{hendrycks2021measuring}, GSM8K~\citep{cobbe2021training}, and GPQA~\citep{rein2024gpqa}.
\textbf{(4) Representation Tasks:} Finally, we assess the zero-shot generalizability of MemOnce's learned memory representations on downstream text retrieval and reranking benchmarks~\citep{nie2026evoembedding}.

\paragraph{Baselines.}
We compare MemOnce against three distinct categories of baselines corresponding to our evaluation tracks:
\textbf{(1) Foundational Architectures:} We compare against the standard Transformer~\citep{vaswani2017attention}, cross-layer KV-sharing variants including YOCO~\citep{sun2024you} and LastKV~\citep{cai2026last}, linear and efficient attention architectures including Sliding Window Attention (SWA)~\citep{beltagy2020longformer}, Gated DeltaNet (GDN)~\citep{yang2025gated}, and Multi-head Latent Attention (MLA)~\citep{liu2024deepseek}, as well as the memory architecture Memory Caching~\citep{behrouz2026memory}.\footnote{We exclude Titans~\citep{behrouz2025titans} from our pretraining comparison as its official implementation is not publicly available, and our reproduction exhibited optimization instabilities.}
\textbf{(2) Native Memory Models:} For continual memory evaluation, we compare against advanced native memory methods, including M+~\citep{wang2025m}, Metis~\citep{zhang2026metis}, $\delta$-Mem~\citep{lei2026delta}, and EvoEmbedding~\citep{nie2026evoembedding}.
\textbf{(3) Retrieval and Reranking Models:} For downstream representation tasks, we benchmark against specialized embedding and reranking models, including BGE-M3~\citep{chen2024bge}, as well as Qwen3-Embedding-4B and Qwen3-Reranker-4B~\citep{zhang2025qwen3}.

\paragraph{Implementation Details.}
For foundational capability evaluation, all architectures are trained from scratch on the FineWeb-Edu dataset~\citep{lozhkov2024fineweb} under unified training configurations and token budgets. 
For long-context memory evaluation, we adopt Qwen3.5~\citep{qwen35blog} as the base backbone for lightweight LoRA post-training adaptation, and employ DeepSeek-V4.1-Flash~\citep{xu2026deepseek} as a strict judge to report final response accuracy across all methods.
Unless otherwise specified, MemOnce is configured with a compression ratio of $r = 8$ and retrieves the top-$K = 64$ memory chunks during sparse attention. 
All experiments are conducted on a single server equipped with 8 NVIDIA H100 GPUs. 
Full details are provided in Appendix~\ref{app:experiments}.

\subsection{Main Results}
\label{subsec:overall_performance}

\paragraph{(1) Foundational Capability.}
\textbf{Joint recurrence and compression of context caches across depth and time enable MemOnce to cut the persistent cache footprint by $Lr\times$ while delivering strong foundational performance.}
We report the results of pretraining different foundation architectures from scratch on FineWeb-Edu. 
As shown in Figure~\ref{fig:encoder_decoder}, MemOnce attains lower final validation perplexity than GDN, SWA, and YOCO, matches or surpasses the Transformer even at $r=8$, and achieves the best scaling performance from 0.15B to 1.0B parameters.
In downstream evaluations (Table~\ref{tab:foundation_main}), $\text{MemOnce}_{r=1}$ achieves the highest average zero-shot accuracy across all evaluated architectures (including the extended baselines in Appendix Table~\ref{tab:full_400m_results}), while $\text{MemOnce}_{r=8}$ ranks second in Table~\ref{tab:foundation_main} and outperforms its underlying GDN baseline (52.35 vs.\ 52.31) and Hybrid Backbone (51.99), with the two variants respectively achieving the lowest language modeling perplexities on WikiText-103 and LAMBADA.

Notably, compressing the temporal dimension by $8\times$ ($\text{MemOnce}_{r=8}$ vs.\ $\text{MemOnce}_{r=1}$) incurs virtually no performance loss, demonstrating that the model can effectively condense segment-level context into the trailing tokens while preserving essential long-range information.
Crucially, whereas linear models (DeltaNet and GDN) degrade in language modeling perplexity due to fixed-state capacity bottlenecks and prior cache-sharing methods (YOCO and LastKV) lag behind in reasoning accuracy due to direct cross-layer reuse without learned decoding, MemOnce excels across both dimensions, validating its expressiveness as a foundation architecture.

\begin{table*}[!t]
\centering
\caption{\textbf{Performance of foundation architectures on language modeling and zero-shot common-sense reasoning tasks.} All models are pretrained from scratch on FineWeb-Edu under fair configurations. {Wiki.} and {LMB.} denote perplexity ($\downarrow$) on WikiText-103 and LAMBADA, respectively. Best results are in \textbf{bold}, and second-best are \underline{underlined}.}
\label{tab:foundation_main}
\resizebox{.99\textwidth}{!}{
\begin{tabular}{lcc|ccccccccc}
\toprule
\textbf{Model} & \textbf{Wiki.} & \textbf{LMB.} & \textbf{LMB.} & \textbf{PIQA} & \textbf{Hella.} & \textbf{Wino.} & \textbf{ARC-e} & \textbf{ARC-c} & \textbf{SciQ} & \textbf{BoolQ} & \textbf{Avg.} \\
 & ppl $\downarrow$ & ppl $\downarrow$ & acc $\uparrow$ & acc $\uparrow$ & acc\_n $\uparrow$ & acc $\uparrow$ & acc\_n $\uparrow$ & acc\_n $\uparrow$ & acc $\uparrow$ & acc $\uparrow$ & $\uparrow$ \\
\midrule
\multicolumn{12}{l}{\textbf{\textit{Transformers \& Attention Variants}}} \\[1pt]
Transformer~\citep{vaswani2017attention} & 20.90 & 33.24 & \underline{35.2} & 65.7 & 40.4 & 51.5 & \underline{52.4} & \textbf{28.2} & 83.2 & 58.2 & 51.85 \\
SWA~\citep{beltagy2020longformer} & 20.52 & 32.69 & 34.5 & 65.6 & 41.0 & 51.4 & 50.1 & 27.4 & 83.2 & 59.7 & 51.60 \\
MLA~\citep{liu2024deepseek} & 23.07 & 33.37 & 34.2 & 67.0 & 39.7 & 50.7 & 50.6 & 27.7 & 83.2 & \textbf{61.9} & 51.87 \\
\midrule
\multicolumn{12}{l}{\textbf{\textit{Linear \& Recurrent Architectures}}} \\[1pt]
DeltaNet~\citep{yang2024parallelizing} & 23.75 & 33.71 & 30.8 & 65.6 & 37.9 & 51.0 & 49.1 & 26.8 & 80.7 & 60.3 & 50.28 \\
GDN~\citep{yang2025gated} & 26.03 & 33.37 & 34.0 & \underline{67.6} & 40.6 & 50.3 & \textbf{54.1} & 27.2 & \underline{84.2} & 60.6 & 52.31 \\
\midrule
\multicolumn{12}{l}{\textbf{\textit{Efficient \& Memory Architectures}}} \\[1pt]
Hybrid & \underline{19.56} & \underline{32.61} & 35.1 & 66.5 & \underline{42.0} & \underline{51.7} & 51.4 & 27.1 & 83.4 & 58.7 & 51.99 \\
YOCO~\citep{sun2024you} & 24.38 & 35.18 & \textbf{36.1} & 67.4 & 40.7 & 51.1 & 51.6 & 26.5 & 82.6 & 59.7 & 51.97 \\
LastKV~\citep{cai2026last} & 21.12 & 34.04 & 33.7 & 65.5 & 40.0 & 50.5 & 50.9 & \underline{27.8} & 81.5 & 58.8 & 51.09 \\
Memory Caching~\citep{behrouz2026memory} & 29.84 & 42.47 & 17.6 & 62.6 & 33.5 & 49.9 & 47.5 & 25.7 & 72.8 & 56.6 & 45.78 \\
\midrule
\rowcolor{rowlight} \textbf{MemOnce$_{r=8}$ (Ours)} & 21.08 & \textbf{32.45} & 34.0 & \textbf{67.8} & 41.9 & \textbf{51.8} & 52.3 & 27.7 & 81.7 & \underline{61.6} & \underline{52.35} \\
\rowcolor{rowdeep} \textbf{MemOnce$_{r=1}$ (Ours)} & \textbf{19.28} & 32.69 & \underline{35.2} & \underline{67.6} & \textbf{42.2} & 49.6 & \underline{52.4} & 27.5 & \textbf{84.7} & \textbf{61.9} & \textbf{52.63} \\
\bottomrule
\end{tabular}
}
\end{table*}

\begin{table*}[!t]
\centering
\caption{\textbf{Performance on five memory benchmarks across varying context lengths.} MemOnce is the first native memory model to surpass the full-context baseline. Best results are in \textbf{bold} and second-best are \underline{underlined}. $^\dagger$ denotes re-trained models for fair comparison.}
\label{tab:memory_main}
\resizebox{.9\textwidth}{!}{
\begin{tabular}{lc|ccccc|c}
\toprule
\multirow{2}{*}{\textbf{Model}} & \multirow{2}{*}{\textbf{Size}} & \textbf{MemBench} & \textbf{LoCoMo} & \textbf{LongMemEval} & \textbf{PersonaMem-128} & \textbf{PersonaMem-1M} & \textbf{Mem. Avg.} \\
& & ($\sim$4K) $\uparrow$ & ($\sim$21K) $\uparrow$ & ($\sim$115K) $\uparrow$ & ($\sim$128K) $\uparrow$ & ($\sim$1M) $\uparrow$ & $\uparrow$ \\
\midrule
\multicolumn{8}{l}{\textbf{\textit{Full Context Baselines}}} \\[1pt]
Qwen3.5-2B & 2B & 72.69 & 64.35 & 41.00 & 45.00 & 23.60 & 49.33 \\
Qwen3.5-4B & 4B & \underline{82.64} & \textbf{69.55} & \underline{42.20} & 43.50 & 33.60 & \underline{54.30} \\
\midrule
\multicolumn{8}{l}{\textbf{\textit{Native Memory Models}}} \\[1pt]
$\delta$-Mem~\citep{lei2026delta} & 4B & 46.72 & 16.88 & 11.20 & 40.00 & 31.60 & 29.28 \\
EvoEmbedding~\citep{nie2026evoembedding} & 4B & 67.08 & 25.97 & 16.40 & 47.30 & 39.00 & 39.15 \\
Metis~\citep{zhang2026metis} & 4B & 47.45 & 9.35 & 7.40 & 37.60 & 37.20 & 27.80 \\
M+~\citep{wang2025m} & 8B & 66.93 & 46.43 & 27.40 & 38.40 & 27.80 & 41.39 \\
\cmidrule(lr){1-8}
$\delta$-Mem$^\dagger$~\citep{lei2026delta} & 4B & 55.53 & 15.00 & 13.00 & 34.00 & 32.20 & 29.95 \\
EvoEmbedding$^\dagger$~\citep{nie2026evoembedding} & 4B & 69.13 & 24.68 & 18.60 & 42.50 & 38.80 & 38.74 \\
Metis$^\dagger$~\citep{zhang2026metis} & 4B & 50.03 & 25.06 & 8.80 & 38.90 & 36.00 & 31.76 \\
\midrule
\rowcolor{rowdeep} \textbf{MemOnce (Ours)} & 2B & 73.54 & 59.03 & 35.40 & \underline{48.50} & \underline{39.40} & 51.17$_{+1.84}$ \\
\rowcolor{rowdeep} \textbf{MemOnce (Ours)} & 4B & \textbf{84.34} & \underline{67.01} & \textbf{47.40} & \textbf{53.60} & \textbf{48.00} & \textbf{60.07}$_{+5.77}$ \\
\bottomrule
\end{tabular}
}
\end{table*}

\begin{table}[!t]
\centering
\caption{\textbf{General reasoning performance under a 32K memory load across different memory paradigms.} Best results are in \textbf{bold} and second-best are \underline{underlined}. Subscripts indicate changes relative to the Qwen3.5-4B base model.}
\label{tab:general_bench}
\resizebox{.7\columnwidth}{!}{
\begin{tabular}{ll|ccc|c}
\toprule
\textbf{Model} & \textbf{Mechanism} & \textbf{MMLU} & \textbf{GSM8K} & \textbf{GPQA} & \textbf{Gen. Avg.} \\
\midrule
Qwen3.5-4B & Full Context & \underline{72.81} & \textbf{92.00} & 32.83 & 65.88 \\
Qwen3.5-4B$_{\text{full sft}}$ & Model Weight & 51.97 & 40.50 & 22.73 & 38.40$_{-27.48}$ \\
\midrule
EvoEmbedding~\citep{nie2026evoembedding} & Soft Prompt & 68.64 & 39.50 & \textbf{34.85} & 47.66$_{-18.22}$ \\
Metis~\citep{zhang2026metis} & Linear State & 71.71 & 23.50 & \textbf{34.85} & 43.35$_{-22.53}$ \\
$\delta$-Mem~\citep{lei2026delta} & Linear State & \textbf{73.46} & \underline{91.00} & 33.33 & \underline{65.93}$_{+0.05}$ \\
\midrule
\rowcolor{rowdeep} \textbf{MemOnce (Ours)} & KV Cache & 72.59 & \underline{91.00} & \underline{34.34} & \textbf{65.98}$_{+0.10}$ \\
\bottomrule
\end{tabular}
}
\end{table}

\paragraph{(2) Long-Context Memory Capability.}
\textbf{MemOnce is the first native memory model to surpass full-context baselines across model and context scales without suffering from catastrophic forgetting.}
Table~\ref{tab:memory_main} reports evaluation results across five long-context memory benchmarks spanning 4K to 1M tokens. 
Across both 2B and 4B scales, MemOnce surpasses its full-context Qwen3.5 counterparts with average scores of $51.17\%$ ($+1.84\%$) and $60.07\%$ ($+5.77\%$), respectively, as query-conditioned sparse retrieval filters out irrelevant historical context and mitigates the attention dilution inherent to ultra-long sequences.
Moreover, MemOnce substantially outperforms both official and re-trained ($\dagger$) native memory baselines, including M+ ($41.39\%$), EvoEmbedding ($39.15\%$), Metis$^\dagger$ ($31.76\%$), and $\delta$-Mem$^\dagger$ ($29.95\%$).
Existing fixed-state memory methods suffer from severe representational bottlenecks when compressing long histories, losing fine-grained details required by multi-session benchmarks such as LoCoMo and LongMemEval. In contrast, our post-training adaptation maintains an expandable state pool and decodes retrieved representations into multi-layer KV caches, preserving the full expressiveness of attention without imposing the capacity limitations of linear models.
Meanwhile, under a 32K memory load on general reasoning tasks (Table~\ref{tab:general_bench}), methods based on soft-prompt injection (EvoEmbedding, $-18.22\%$) and in-backbone linear states (Metis, $-22.53\%$) perturb internal model activations and induce severe catastrophic forgetting, while $\delta$-Mem preserves general accuracy ($65.93\%$) at the cost of poor memory recall ($29.28\%$).
By keeping backbone weights frozen during generation and internalizing history strictly as decoded KV caches, MemOnce achieves the highest general reasoning average alongside superior memory recall.

\begin{figure*}[!t]
\centering
\includegraphics[width=.99\textwidth]{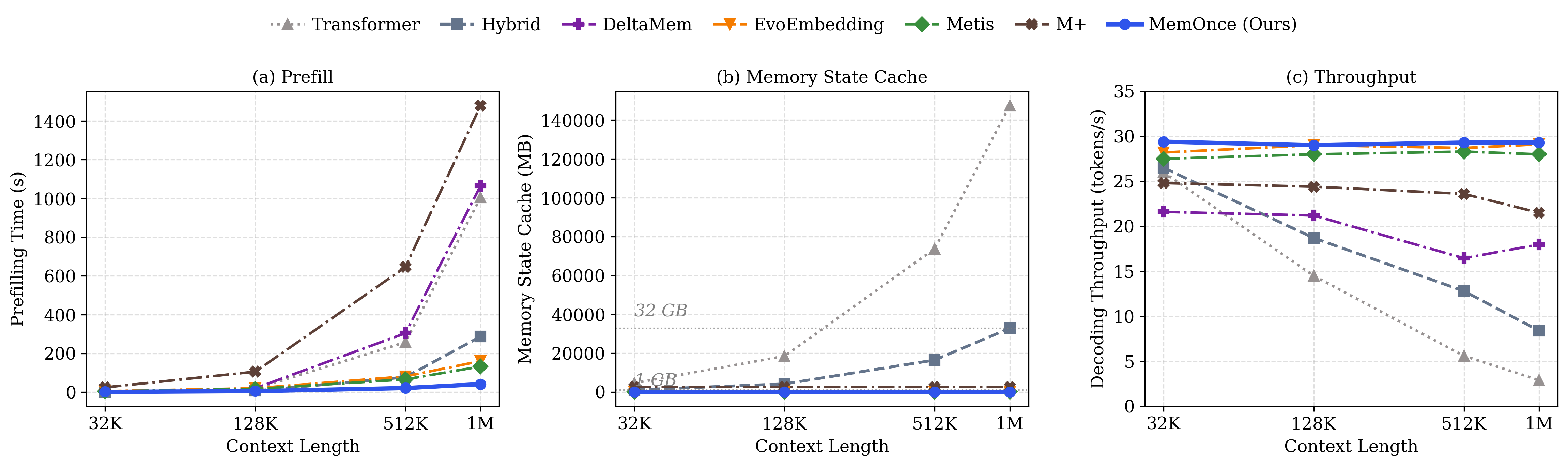}
\caption{\textbf{Inference scalability across context lengths from 32K to 1M tokens.}}
\label{fig:memory_efficiency_curves}
\end{figure*}

\paragraph{(3) Inference Efficiency and Scalability.}
\textbf{MemOnce achieves linear-time prefilling, constant active GPU cache, and steady decoding throughput up to 1M tokens.}
Figure~\ref{fig:memory_efficiency_curves} evaluates prefilling time, GPU cache footprint, and decoding throughput across context lengths from 32K to 1M tokens (detailed in Table~\ref{tab:efficiency_scaling}).
Beyond the quadratic prefill cost of standard Transformers, existing memory baselines also suffer from severe execution bottlenecks: $\delta$-Mem still relies on full-context attention computation during state construction in its official implementation, Metis employs coarse sequential blocks that limit GPU parallelism, EvoEmbedding requires two separate forward passes to generate memory states and retrieval keys, and M+ incurs heavy layer-wise retrieval and CPU--GPU communication overhead across all layers.
By encoding historical chunks in parallel and retrieving memory only once, MemOnce scales linearly from 1.38s (32K) to 41.36s (1M), achieving a $6.9\times$ speedup over Hybrid and $35.8\times$ over M+.
Meanwhile, whereas the KV caches of Transformer and Hybrid grow linearly with sequence length and exceed $32\text{ GB}$ at 1M tokens, MemOnce offloads the historical pool and bounds the active GPU cache to a near-constant $25\text{ MB}$ footprint across all scales (Figure~\ref{fig:memory_efficiency_curves}(b)).
Consequently, while surging KV cache traffic severely degrades decoding throughput in Hybrid and Transformer baselines, MemOnce attends only to a constant-sized memory subset, sustaining the highest and near-invariant throughput across all context lengths.

\begin{table*}[!t]
\centering
\caption{\textbf{Retrieval performance across diverse long-context and conversational benchmarks.} We report Recall@10 (R@10) and NDCG@10 (N@10). Best results within each metric block are in \textbf{bold} and second-best are \underline{underlined}.}
\label{tab:retrieval_bench}
\resizebox{\textwidth}{!}{
\begin{tabular}{lccc|cccccccc|c}
\toprule
\textbf{Model} & \textbf{Size} & \textbf{Dim} & \textbf{Metric} & \textbf{ESG-Reports} & \textbf{LoCoMo} & \textbf{LongMemEval} & \textbf{REALTALK} & \textbf{QASPER} & \textbf{PeerQA} & \textbf{CovidQA} & \textbf{MLDR} & \textbf{Overall} \\
\midrule
BGE-M3 & 1.2B & 1024 & R@10 & 51.00 & \textbf{55.70} & 77.80 & \textbf{52.40} & 66.40 & \textbf{37.30} & 87.20 & \underline{97.00} & \underline{65.60} \\
Qwen3-Embedding-4B & 4B & 2560 & R@10 & 57.90 & 41.30 & 58.70 & 42.70 & \underline{72.90} & \underline{36.00} & \textbf{93.20} & 96.00 & 62.30 \\
\rowcolor{rowdeep} \textbf{MemOnce (Ours)} & 4B & 1024 & R@10 & \textbf{72.55} & \underline{52.54} & \textbf{86.18} & \underline{44.14} & \textbf{76.47} & 35.38 & \underline{88.39} & \textbf{98.00} & \textbf{69.21} \\
\midrule
\midrule
BGE-M3 & 1.2B & 1024 & N@10 & 35.90 & \textbf{41.80} & 65.60 & \textbf{42.00} & 44.20 & \textbf{28.20} & \underline{65.10} & 77.00 & \underline{50.00} \\
Qwen3-Embedding-4B & 4B & 2560 & N@10 & 40.80 & 29.90 & 47.10 & \underline{34.50} & \underline{50.20} & \underline{27.20} & \textbf{76.70} & 77.80 & 48.10 \\
\rowcolor{rowdeep} \textbf{MemOnce (Ours)} & 4B & 1024 & N@10 & \textbf{55.82} & \underline{38.68} & \textbf{72.75} & 31.78 & \textbf{53.64} & 25.05 & 64.68 & \textbf{82.09} & \textbf{53.06} \\
\bottomrule
\end{tabular}
}
\end{table*}

\begin{table}[!t]
\centering
\caption{\textbf{Context reranking performance across agent memory systems on the LoCoMo benchmark.} All methods use All-MiniLM-L6-v2 to retrieve 256 candidate segments and restrict the final generator context to Top-20. Best results within each system are in \textbf{bold} and second-best are \underline{underlined}.}
\label{tab:rerank_locomo}
\resizebox{.7\columnwidth}{!}{
\begin{tabular}{ll|cccc|c}
\toprule
\textbf{System} & \textbf{Method} & \textbf{Single-hop} & \textbf{Multi-hop} & \textbf{Temporal} & \textbf{Open} & \textbf{Overall} \\
\midrule
\multirow{4}{*}{\textbf{A-MEM}} 
 & Original & 42.91 & 13.71 & 32.29 & 56.48 & 43.57 \\
 & + Reasoning & 43.26 & 15.26 & 35.42 & 58.98 & 45.52$_{+2.0}$ \\
 & + Qwen3-Reranker-4B & \underline{61.70} & \underline{22.12} & \underline{43.75} & \textbf{76.46} & \underline{60.39}$_{+16.8}$ \\
 & \cellcolor{rowdeep}\textbf{+ MemOnce (Ours)} & \cellcolor{rowdeep}\textbf{66.31} & \cellcolor{rowdeep}\textbf{33.96} & \cellcolor{rowdeep}\textbf{46.88} & \cellcolor{rowdeep}\underline{70.75} & \cellcolor{rowdeep}\textbf{60.78}$_{+17.2}$ \\
\midrule
\multirow{4}{*}{\textbf{LightMem}} 
 & Original & 41.13 & 23.36 & \underline{41.67} & 46.85 & 40.58 \\
 & + Reasoning & 48.23 & 24.92 & 39.58 & 49.94 & 43.77$_{+3.2}$ \\
 & + Qwen3-Reranker-4B & \underline{55.32} & \textbf{29.28} & \textbf{44.79} & \textbf{61.12} & \underline{52.40}$_{+11.8}$ \\
 & \cellcolor{rowdeep}\textbf{+ MemOnce (Ours)} & \cellcolor{rowdeep}\textbf{67.02} & \cellcolor{rowdeep}\underline{27.10} & \cellcolor{rowdeep}\underline{41.67} & \cellcolor{rowdeep}\textbf{61.12} & \cellcolor{rowdeep}\textbf{53.90}$_{+13.3}$ \\
\bottomrule
\end{tabular}
}
\end{table}

\paragraph{(4) Representation Capability.}
\textbf{The memory representations learned by MemOnce transfer naturally to downstream retrieval and context reranking tasks, outperforming specialized embedding and reranking models.}
We evaluate MemOnce on eight long-context and conversational retrieval benchmarks against dedicated embedding models, BGE-M3 and Qwen3-Embedding-4B~\citep{zhang2025qwen3} (Table~\ref{tab:retrieval_bench}). 
Unlike standalone embedding models trained solely on pairwise text similarity, MemOnce jointly optimizes its terminal index representations with end-to-end language modeling and contrastive evidence supervision, enabling the compact 1024-dimensional descriptors to capture multi-hop and temporal dependencies critical for answering downstream queries.
Consequently, MemOnce achieves the best overall Recall@10 of $69.21$ and NDCG@10 of $53.06$, ranking first on long-context benchmarks. 
Furthermore, when deployed as a plug-and-play reranker on LoCoMo to select the top-20 segments from 256 candidates retrieved by All-MiniLM-L6-v2~\citep{nie2026evoembedding} (Table~\ref{tab:rerank_locomo}), MemOnce boosts the overall performance of representative agent memory systems by $+17.2$ on A-MEM ($60.78$) and $+13.3$ on LightMem ($53.90$), consistently surpassing both the multi-turn reasoning-based retrieval strategy and the specialized Qwen3-Reranker-4B~\citep{zhang2025qwen3}.

\paragraph{Ablation Summary.}
In Appendix~\ref{sec:ablation}, we further ablate the core components of MemOnce and observe that: (1) increasing the compression ratio from $r=1$ to $r=8$ and discarding cross-segment linear states incur less than $0.3\%$ accuracy loss, demonstrating that final-layer hidden states serve as a highly effective carrier for compressed memory without relying on recurrent linear states (Table~\ref{tab:ablation_compression}); (2) retrieving $K=64$ memory chunks strikes the optimal balance between recall coverage and contextual noise (Table~\ref{tab:ablation_topk}); and (3) single-shot indexing reduces query-time retrieval latency by $12.0\times$ over M+ and $17.0\times$ over RAG-Rewrite (Table~\ref{tab:ablation_efficiency}). Additional scaling and RAG comparisons are provided in Appendix~\ref{app:results}.

\section{Discussion}
\label{sec:discussion}

\subsection{Why Existing Memory Architectures Fail to Scale}
Existing memory architectures~\citep{wang2025m,lei2026delta,zhang2026metis} struggle to scale effectively due to three fundamental bottlenecks: performance degradation, system overhead, and limited generalizability. 
First, from a performance perspective, compressing full-context attention into bounded or fixed-size states inevitably induces information decay and fails to retain fine-grained details over long horizons.
Second, from a cost perspective, long-term memory methods such as M+~\citep{wang2025m} overlook cache compression, as maintaining multi-layer KV caches and repeatedly fetching states across layers incur heavy storage overhead and communication latency.
Third, from a generalizability perspective, prior memory architectures are typically tailored either to pretraining from scratch~\citep{behrouz2025titans} or to post-training adaptation, lacking versatility across both paradigms.
To tackle these challenges, MemOnce couples compact final-layer memory encoding with efficient memory retrieval and on-demand KV decoding, and is validated across both pretraining and post-training adaptation.
Our experimental results demonstrate that learning scalable, reusable memory states provides a viable path toward next-generation foundation architectures.

\subsection{Limitations and Future Work}
Several limitations of MemOnce offer valuable directions for future research. 
First, constrained by available computing resources, our pre-training from scratch is conducted up to the 1B parameter scale; validating the foundational scaling behavior of MemOnce via full pre-training at larger scales (e.g., 7B or beyond) remains future work. 
Second, while MemOnce demonstrates strong memory capabilities, its performance in downstream agent tasks remains to be fully validated, particularly in complex scenarios such as long-horizon interactive coding and multi-step tool use. 
In future work, we plan to pre-train MemOnce at larger parameter scales and explore its potential across broader continual interactive scenarios.

%% file: 6_conclusion.tex
\section{Conclusion}
\label{sec:conclusion}

In this paper, we present \textbf{MemOnce}, a foundational memory architecture that achieves accurate long-context recall, linear-time complexity, and constant active memory states for continual language modeling. 
By coupling the memory encoder-decoder and memory sparse attention designs, MemOnce caches and retrieves memory states only once while dynamically decoding them into multi-layer KV caches on demand.
MemOnce seamlessly integrates into pretrained models to accelerate inference while establishing itself as the first native memory model to surpass full-context baselines. 
Extensive experiments across diverse training paradigms, model scales, and context lengths demonstrate its effectiveness and generality.

%% file: appendix.tex
\onecolumn
\setcounter{page}{1}
\begin{center}
  {\Large\textbf{\thetitle}}\\[0.5em]
  {\Large Supplementary Material}\\[1.0em]
\end{center}

\begingroup
  \makeatletter
  \def\@part[#1]#2{\refstepcounter{part}\addcontentsline{toc}{part}{#2}}%
  \makeatother
  \part{Supplementary Material}%
\endgroup
\begingroup
  \hypersetup{linkcolor=selfevolagent_dark}%
  \etocsettocstyle{\section*{Contents}}{}%
  \etocsetnexttocdepth{2}%
  \makeatletter
  \renewcommand*\l@section[2]{%
    \ifnum \c@tocdepth >\z@
      \addpenalty\@secpenalty
      \addvspace{0.4em}%
      \setlength\@tempdima{1.5em}%
      \begingroup
        \parindent \z@ \rightskip \@pnumwidth
        \parfillskip -\@pnumwidth
        \leavevmode \bfseries
        \advance\leftskip\@tempdima
        \hskip -\leftskip
        #1\nobreak\hfil\nobreak\hb@xt@\@pnumwidth{\hss #2}\par
      \endgroup
    \fi}
  \makeatother
  \localtableofcontents
\endgroup

\setcounter{section}{0}
\renewcommand{\thesection}{\Alph{section}}

\section{Related Work}
\label{sec:related_work}

\subsection{Architectural and Attention Optimizations}
To alleviate the quadratic compute and linear KV cache growth of standard Transformers~\citep{vaswani2017attention}, architectural reforms have progressed along three lines. 
First, \textit{from linear to hybrid architectures}: linear and recurrent models such as Mamba~\citep{gu2023mamba}, RWKV~\citep{peng2023rwkv}, and Gated DeltaNet~\citep{yang2025gated} compress history into fixed-size states for constant-memory, linear-time inference, but struggle with fine-grained in-context recall~\citep{jelassi2024repeat}. 
To balance computational efficiency with accurate recall of historical information, modern backbones increasingly adopt hybrid designs: Qwen3.5~\citep{qwen35blog} and Kimi K3~\citep{team2026kimi} interleave linear or gated delta layers with periodic global attention, while MiMo~\citep{xiaomi2026mimo} interleaves sliding-window attention with full attention. 
Second, \textit{from global to sparse attention}: intra-layer sparse mechanisms, such as Native Sparse Attention (NSA)~\citep{yuan2025native,xu2026deepseek} and block-sparse routing~\citep{qiu2026design}, dynamically prune attention matrices via hardware-aligned compressed, selected, and sliding-window branches. 
Third, \textit{cross-layer KV cache reuse}: architectures like YOCO~\citep{sun2024you} and HySparse~\citep{hysparse2026} cache KV states at an intermediate anchor layer and feed them forward to subsequent layers, whereas LastKV~\citep{cai2026last} directly reuses raw final-layer KV states across earlier layers.
Despite these gains, hybrid and sparse models still retain multi-layer KV caches, with sparse mechanisms further incurring repeated routing overhead across layers and decoding steps, while cross-layer sharing methods still retain full-length token states without sequence compression.

\subsection{Native Memory Models}
Native memory models internalize historical context into compact parametric or latent states. 
One category compresses history into fixed-capacity representations: Titans~\citep{behrouz2025titans} updates a deep neural memory at test time via surprise-driven gradients with momentum and weight decay; $\delta$-Mem~\citep{lei2026delta} updates low-dimensional state matrices online via delta-rule learning; Metis~\citep{zhang2026metis} embeds persistent, gradient-free memory states directly into the backbone in a single forward pass; and EvoEmbedding~\citep{nie2026evoembedding} maintains a window-based latent memory queue for long-context representation learning. 
To overcome fixed capacity limits, a second category dynamically expands the state pool: M+~\citep{wang2025m} extends MemoryLLM~\citep{wang2024memoryllm} with a long-term latent pool and layer-wise co-trained retrievers; and Memory Caching~\citep{behrouz2026memory} caches recurrent states at segment boundaries for gated residual aggregation.
Nevertheless, these methods either suffer from fixed-capacity bottlenecks or incur retrieval overhead, and their prevalent reliance on post-training adaptation often induces catastrophic forgetting.

\subsection{External Memory Systems}
External memory systems decouple persistent memory storage and management from the model backbone, representing historical experiences in explicit stores and handling memory construction, updating, and retrieval through dedicated pipelines~\citep{hu2025memory}.
Mem0~\citep{chhikara2025mem0} dynamically extracts and consolidates salient conversational facts for subsequent retrieval, while A-MEM~\citep{xu2026mem} employs agent-driven note construction and dynamic linking to organize memories into an evolving knowledge network.
For multimodal personalization, PersonaVLM~\citep{nie2026personavlm} integrates proactive multimodal memory accumulation and multi-turn memory-grounded reasoning with dynamic personality profiling for response alignment.
To improve memory construction efficiency, LightMem~\citep{fang2026lightmem} introduces lightweight sensory-level compression, topic-aware segmentation and summarization, and offline ``sleep-time'' consolidation.
For streaming video understanding, Light-Omni~\citep{nie2026lightomni} employs a dual-state architecture, combining a non-parametric global state distilled from episodic memory with a parametric latent state that enables reflexive action control and semantically aligned retrieval without iterative reasoning.
Despite these advances, long-term memories are still largely maintained in external stores or auxiliary memory structures, with memory construction, consolidation, and retrieval handled through separate procedures. This separation can introduce additional storage and orchestration overhead while making it less straightforward to jointly optimize memory formation, maintenance, and utilization with the backbone, particularly under end-to-end objectives such as reinforcement learning~\citep{wang2025memalpha}.

\subsection{Comparison with Existing Methods}
MemOnce addresses the limitations across all three paradigms.
Compared with architectural and cross-layer KV optimizations, MemOnce caches only a small subset of final-layer hidden states and performs sparse retrieval only once per query, eliminating both layer-wise storage and routing overhead.
Compared with native and external memory systems, MemOnce enables efficient end-to-end optimization and long-term memory capabilities for both pretraining and post-training adaptation.
Furthermore, decoding compact memory states into KV caches effectively preserves long-range context without catastrophic forgetting.

\section{Detailed Analysis and Implementation of MemOnce}
\label{app:memonce_analysis}

\subsection{Test-Time Memorization and Retrieval}
\label{app:subsec_multiblock_index}

\paragraph{Parallel Encoding and Two-Tier Storage ($G_t$).}
To accelerate memory construction during prefill, MemOnce processes historical inputs $x_{1:t-1}$ in parallel chunks of length $n_i$ (e.g., $n_i = 128$ in pretraining and $n_i = 512$ in post-training adaptation) under a compression ratio of $r = 8$. 
For each chunk $x_i$, MemOnce retains its trailing $n_i / r$ final-layer token states $m_i = h_{i, \,-n_i/r:}^L \in \mathbb{R}^{(n_i/r) \times D}$ to form the persistent memory bank $G_t = [m_1; \dots; m_B] \in \mathbb{R}^{k \times D}$ across $B$ chunks ($k = \lfloor N/r \rfloor$).
Because MemOnce resets linear-block recurrent states across chunk boundaries (validated in Table~\ref{tab:ablation_compression}), caches only final-layer states, and requires only short-term memory without long-term retrieval during encoding, historical inputs $x_{1:t-1}$ can be encoded in parallel batches without sequential dependencies.
As defined in Section~\ref{subsec:sparse_attention}, $G_t$ is organized into a two-tier storage hierarchy: 
(1) the \textbf{short-term memory window} $G_t^{\text{short}}$, which maintains the most recent compressed states (e.g., $1024$ compressed tokens across $16$ chunks at $n_i=512, r=8$, spanning $8\text{K}$ raw tokens) as a sliding window on the GPU to preserve local continuity with zero retrieval latency; and 
(2) the \textbf{long-term memory pool} $G_t^{\text{long}}$, which offloads older states to host memory or SSD once they slide out of $G_t^{\text{short}}$.

\paragraph{Memory Footprint and Bandwidth Analysis.}
On the Qwen3.5-4B hybrid backbone ($D = 2560$, $n_i = 512$, $r = 8$), each compressed 64-token chunk occupies only $320\text{ KB}$. Consequently, the GPU-resident short-term window $G_t^{\text{short}}$ ($16$ chunks) requires a constant footprint of $5.0\text{ MB}$. During retrieval, fetching the top-$K$ ($K=64$) long-term chunks $M_t^{\text{long}}$ from SSD/CPU transfers only $20.0\text{ MB}$ over PCIe (Table~\ref{tab:ablation_efficiency}), yielding a total active memory state $M_t = [M_t^{\text{long}} \,\Vert{}\, G_t^{\text{short}}]$ of just $25.0\text{ MB}$ (vs.\ $1,150.0\text{ MB}$ for 32-layer retrieval in M+).
Once loaded onto the GPU, $M_t$ is decoded on demand into multi-layer KV caches $C_t = \{(K^l, V^l)\}_{l \in \mathcal{L}_{\text{attn}}}$ for the full-attention layers $\mathcal{L}_{\text{attn}} \subseteq \{1, \dots, L\}$ via $u_\theta$, where each layer-specific projection $K_{\text{Proj}}^l$ and $V_{\text{Proj}}^l$ consists of LayerNorm, a linear projection, and key normalization, leaving all linear GDN blocks free of historical cache overhead.

\paragraph{Indexing Projection and Memory State Routing.}
To index $G_t^{\text{long}}$ without external embedding models or layer-wise routers, the projection $\phi: \mathbb{R}^{D} \to \mathbb{R}^{d_e}$ ($d_e = 1024$) in Eq.~\eqref{eq:sparse_retrieval} extracts the terminal token state $h_{i, -1}^L \in \mathbb{R}^D$ of each compressed chunk $m_i$ (denoted as $\phi(m_i) \triangleq \phi(h_{i, -1}^L)$) and maps it via a two-layer MLP with SiLU activation followed by $\ell_2$-normalization.
Because causal masking and linear blocks aggregate sequence-level semantics into the terminal token, applying $\phi(\cdot)$ solely to $h_{i, -1}^L$ yields an expressive descriptor without sequence-pooling overhead.
Each index vector $\phi(h_i^L) \in \mathbb{R}^{1024}$ takes only $2\text{ KB}$. Thus, across a 128K context ($256$ chunks), the entire index matrix occupies just $512\text{ KB}$ ($4.0\text{ MB}$ at 1M tokens) and remains permanently resident in GPU memory.
At query time, MemOnce encodes the continual input $x_t$ conditioned on $G_t^{\text{short}}$ to compute the query vector $\phi(h_t^L)$ in $0.25\text{ ms}$, performs a matrix-vector inner product to select the top-$K$ states from $G_t^{\text{long}}$, and sorts the retrieved states in chronological order before concatenation in Eq.~\eqref{eq:active_memory_subset} to preserve temporal causality.

\begin{algorithm}[!t]
\caption{Continual Inference of MemOnce}
\label{alg:app_memonce_impl}
\begin{algorithmic}[1]
\Require Historical context $x_{1:t-1}$, continual input $x_t$, compression ratio $r$, retrieval budget $K$.
\Ensure Generated response $\hat{y}_t$.

\Statex \textcolor{gray}{\textit{// Step 1: Memory Writing ($g_\theta$)}}
\State Initialize short-term window $G_t^{\text{short}}$ and long-term pool $G_t^{\text{long}}$.
\For{each historical chunk $x_i \in x_{1:t-1}$}
    \State Extract trailing states $m_i \leftarrow h_{i,\,-n_i/r:}^L$ and index vector $\phi(h_i^L)$.
    \State Update $G_t^{\text{short}} \leftarrow \operatorname{Slide}(G_t^{\text{short}}, m_i)$; offload evicted representations to $G_t^{\text{long}}$.
\EndFor

\Statex \textcolor{gray}{\textit{// Step 2: Memory Indexing ($i_\theta$)}}
\State Compute query representation $\phi(h_t^L)$ conditioned on $G_t^{\text{short}}$.
\State Retrieve top-$K$ states: $M_t^{\text{long}} \leftarrow \operatorname{Top-K}_{m_i \in G_t^{\text{long}}}\, \operatorname{Sim}\big(\phi(h_t^L), \phi(m_i)\big)$ (chronologically sorted).
\State Construct active memory state: $M_t \leftarrow \big[ M_t^{\text{long}} \,\|\, G_t^{\text{short}} \big]$.

\Statex \textcolor{gray}{\textit{// Step 3: On-Demand Decoding ($u_\theta$) and Generation}}
\State Disable memory adapter; decode layer-specific KV caches: $\{K^l, V^l\}_{l=1}^L = u_\theta(M_t)$.
\State Generate response $\hat{y}_t$ via frozen backbone with memory-conditioned sparse attention (Eq.~\eqref{eq:memory_attn}).
\State \Return $\hat{y}_t$
\end{algorithmic}
\end{algorithm}

\subsection{Implementation Workflow}
\label{app:subsec_workflow}

Algorithm~\ref{alg:app_memonce_impl} summarizes the continual inference pipeline of MemOnce. To ensure stable and efficient optimization over long contexts, we adopt three key strategies during post-training adaptation:

\begin{enumerate}
    \item \textbf{Chunked Memory Encoding:} Historical inputs $x_{1:t-1}$ are partitioned into chunks (up to $8$ segments or $4096$ tokens) for parallel encoding. Encoding is performed independently per chunk using the short-term sliding window $G_t^{\text{short}}$ without querying the long-term pool $G_t^{\text{long}}$, eliminating sequential retrieval dependencies during prefill. Training memory is bounded via activation checkpointing and a dedicated LoRA adapter.
    
    \item \textbf{Contrastive Retrieval Supervision:} To align memory routing with language modeling utility, the indexing projection $\phi(\cdot)$ is jointly optimized via an InfoNCE contrastive loss over relevant historical states $m^+ \in G_t^{\text{long}}$:
    \begin{equation}
        \mathcal{L} = -\log \pi_\theta(y_t \mid x_t, M_t) - \lambda \log \frac{\exp\!\big(\operatorname{Sim}(\phi(h_t^L), \phi(m^+)) / \tau\big)}{\sum_{m_i \in G_t^{\text{long}}} \exp\!\big(\operatorname{Sim}(\phi(h_t^L), \phi(m_i)) / \tau\big)},
        \label{eq:training_loss}
    \end{equation}
    where $\tau$ is the temperature hyperparameter and $\lambda$ balances retrieval supervision.
    
    \item \textbf{Frozen-Backbone Generation:} Upon constructing the active memory state $M_t = [M_t^{\text{long}} \,\|\, G_t^{\text{short}}]$, the memory adapter is disabled. The frozen backbone then generates $y_t$ conditioned on the on-demand decoded KV caches $C_t = u_\theta(M_t)$, preserving the base model's general capabilities without catastrophic forgetting.
\end{enumerate}

\section{Training Dataset Construction and Statistics}
\label{app:dataset}

\subsection{MemOnce-1M}
To jointly optimize MemOnce's memory encoder-decoder and single-shot sparse retrieval, we construct a large-scale multi-turn training corpus, \textbf{MemOnce-1M} (1M samples). Every training instance is standardized into a multi-turn conversational sequence $\mathcal{H} = \{x_1, x_2, \dots, x_{t-1}, x_t\}$, consisting of $h = t - 1$ historical interaction rounds followed by a target query turn $x_t$. 
To obtain high-fidelity supervision for both language modeling and memory retrieval without costly manual annotation, we employ \textbf{Qwen3.5-35B-A3B} as a teacher model for the final user query $x_t$. Conditioned on the complete dialogue history, the teacher model simultaneously generates: (i) a grounded reference response $y_t$ to supervise the language modeling objective, and (ii) the exact indices of historical evidence turns $\mathcal{E}^+ \subset \{1, \dots, h\}$ containing the supporting facts required to answer the query. These distilled evidence annotations directly supervise the contrastive retrieval loss in Eq.~\eqref{eq:training_loss}, aligning the query descriptor $\phi(h_t^L)$ with relevant historical index vectors $\{\phi(h_i^L)\}_{i=1}^h$.

\subsection{Data Composition}
As illustrated in Figure~\ref{fig:dataset_stats}(c), our training corpus integrates eight complementary data sources to balance long-range factual recall, multi-hop reasoning, and conversational fluency:
\begin{itemize}
    \item \textbf{Long-Context Representation and Memory Corpora (64.2\%):} The largest portion of the dataset is derived from EvoTrain-180K~\citep{nie2026evoembedding} (\textbf{38.5\%}) and LiveMem\footnote{\url{https://huggingface.co/datasets/chen-l/LiveMem-SFT}} (\textbf{25.7\%}), which provide rich multi-session dialogues and dense long-range retrieval supervision across diverse domains.
    \item \textbf{Multi-Hop and Long-Context Instruction Datasets (25.0\%):} To strengthen complex reasoning over retrieved memory states, we incorporate ChatQA-2\footnote{\url{https://huggingface.co/datasets/nvidia/ChatQA2-Long-SFT-data}} (\textbf{11.0\%}), LongMIT-SFT\footnote{\url{https://github.com/WowCZ/LongMIT}} (\textbf{6.0\%}), 2WikiMultiHopQA\footnote{\url{https://github.com/Alab-NII/2wikimultihop}} (\textbf{5.0\%}), Step-3.5-Flash-SFT\footnote{\url{https://huggingface.co/datasets/stepfun-ai/Step-3.5-Flash-SFT}} (\textbf{2.8\%}), and $\delta$-Mem Qasper~\citep{lei2026delta}\footnote{\url{https://huggingface.co/datasets/huaXiaKyrie/delta-mem-qasper-data}} (\textbf{0.2\%}). Raw long-document QA pairs are segmented into sequential conversational turns so that supporting paragraphs are distributed across separate historical segments.
    \item \textbf{Synthetic Data (10.8\%):} Following the data synthesis methodology of EvoEmbedding~\citep{nie2026evoembedding}, we construct an additional \textbf{10.8\%} of synthetic multi-turn samples (\textbf{Synthetic (Ours)}) via structured template-based question generation. Specifically, we plant multi-source factual chains, evolving user states, and entity attributes across distant dialogue turns using controllable templates, and interleave semantically similar hard-negative distractor turns in between. Positional shortcuts (e.g., explicit turn-index references) are systematically stripped during synthesis, forcing the indexing module and backbone to rely strictly on semantic matching and cross-segment reasoning.
\end{itemize}

\subsection{Statistical Analysis}
\begin{figure*}[!t]
\centering
\begin{subfigure}[b]{0.33\textwidth}
    \centering
    \includegraphics[width=\textwidth]{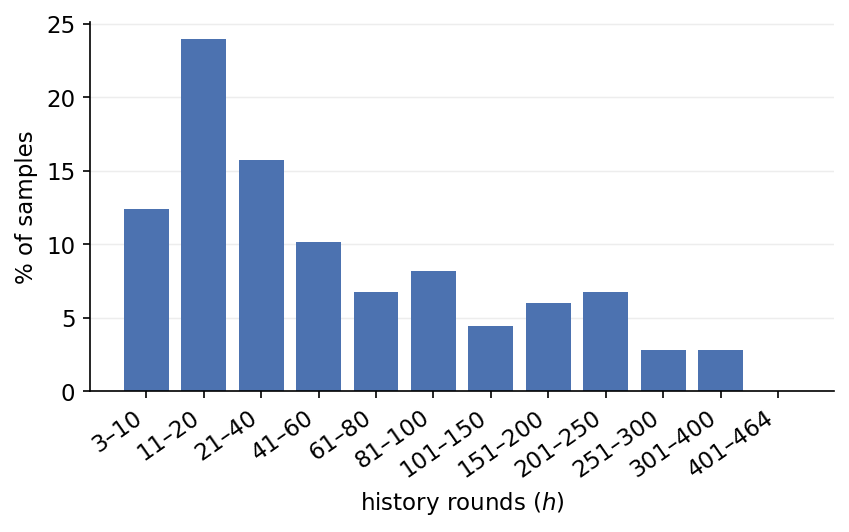}
    \caption{Distribution of history rounds ($h$).}
    \label{fig:dist_rounds}
\end{subfigure}
\hfill
\begin{subfigure}[b]{0.33\textwidth}
    \centering
    \includegraphics[width=\textwidth]{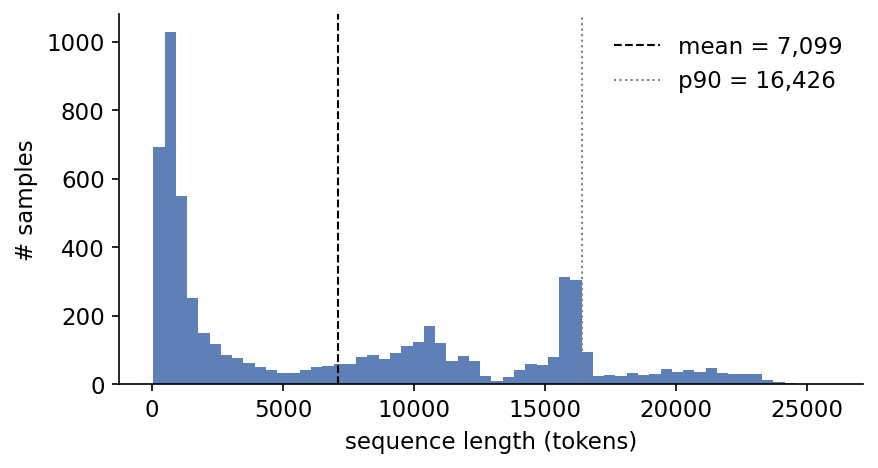}
    \caption{Distribution of sequence lengths.}
    \label{fig:dist_length}
\end{subfigure}
\hfill
\begin{subfigure}[b]{0.31\textwidth}
    \centering
    \includegraphics[width=\textwidth]{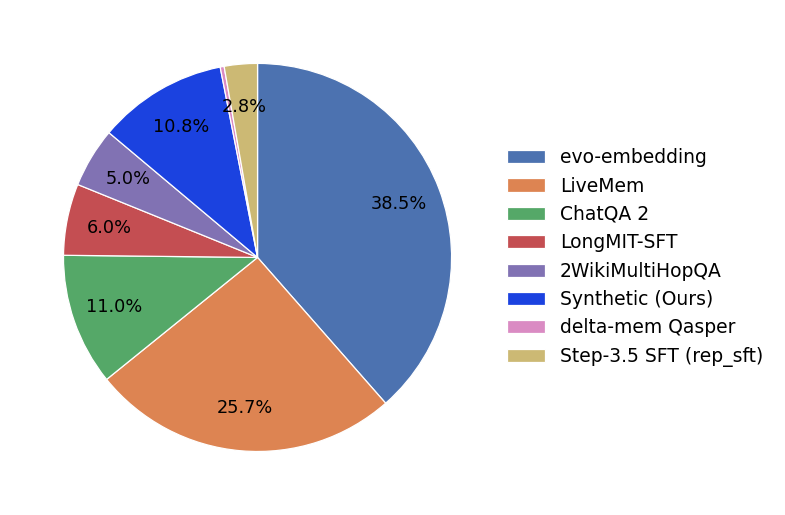}
    \caption{Data source composition.}
    \label{fig:dist_sources}
\end{subfigure}
\caption{\textbf{Statistics of MemOnce-1M.} \textbf{(a)} Percentage of samples across historical dialogue rounds $h \in [3, 464]$. \textbf{(b)} Token sequence length distribution with a mean of $7,099$ tokens and a 90th percentile (\texttt{p90}) of $16,426$ tokens. \textbf{(c)} Proportion of the eight constituent data sources.}
\label{fig:dataset_stats}
\end{figure*}

Figures~\ref{fig:dataset_stats}(a) and~\ref{fig:dataset_stats}(b) present the structural statistics of the dataset across historical conversation rounds and token lengths:
\begin{itemize}
    \item \textbf{Historical Rounds ($h$):} As shown in Figure~\ref{fig:dataset_stats}(a), the number of historical rounds $h$ spans from $3$ to $464$ turns. While short-to-medium dialogues ($3\text{--}10$ and $11\text{--}20$ rounds) account for $12.4\%$ and $24.0\%$ of the corpus (activating a full-window supervision fallback to stabilize early $\phi(\cdot)$ training), over $63\%$ of the samples exceed MemOnce's $16$-segment short-term sliding window $G_t^{\text{short}}$. Specifically, medium-horizon dialogues ($21\text{--}60$ rounds) constitute $25.8\%$, long-horizon sessions ($61\text{--}200$ rounds) account for $25.4\%$, and ultra-long multi-session histories ($201\text{--}464$ rounds) make up the remaining $12.4\%$. This heavy-tailed distribution ensures that the model is rigorously trained to retrieve evicted memory representations from the long-term pool $G_t^{\text{long}}$.
    \item \textbf{Context Lengths:} Figure~\ref{fig:dataset_stats}(b) depicts the token length distribution across dialogue trajectories, exhibiting a multi-modal long-tail profile up to $25,000$ tokens, with a mean of \textbf{7,099 tokens}, a 90th-percentile (\texttt{p90}) of \textbf{16,426 tokens}, and high-density clusters around $10\text{K--}12\text{K}$ and $16\text{K}$ tokens. Training across these multi-chunk trajectories (exceeding our $4096$-token chunk budget) enables MemOnce to generalize seamlessly to $128\text{K}$ and $1\text{M}$ evaluation contexts at test time.
\end{itemize}

\section{Experimental Details}
\label{app:experiments}

\subsection{Training and Implementation Details}
\label{subsec:training_details}

\begin{table}[!t]
\centering
\caption{\textbf{Hyperparameters for pre-training and post-training.}}
\label{tab:training_config}
\small
\begin{tabular}{l|cc}
\toprule
\textbf{Configuration} & \textbf{Pre-training} & \textbf{Post-training} \\
\midrule
Backbone & MemOnce & Qwen3.5 \\
Training Paradigm & Scratch & LoRA (rank$=128, \alpha=256$) \\
Training Dataset & FineWeb-Edu-10B & MemOnce-1M \\
Max Sequence Length & $4$K & $64$K \\
Block Size & $128$ & $512$ \\
Parallel Encoding Chunks & $8$ & $8$ \\
Global Batch Size & $128$ & $64$ \\
Step Data Volume & $0.5$M & --- \\
Training Steps & $21,748$ & $15,625$ \\
Learning Rate & $6\times 10^{-4}$ & $5\times 10^{-5}$ \\
\bottomrule
\end{tabular}
\end{table}

\paragraph{Pre-training Details.}
To evaluate architectural design choices from scratch, we pre-train our base models on FineWeb-Edu-10B~\citep{lozhkov2024fineweb} for one full epoch, tokenized with the Llama-2 BPE tokenizer~\citep{touvron2023llama} (vocabulary size $32{,}000$). All models are optimized with fused AdamW under a $2{,}000$-step linear warmup followed by cosine decay, on $8\times$ NVIDIA H100 GPUs. The full configurations are summarized in Table~\ref{tab:training_config}.

\paragraph{Post-training and Memory Adaptation.}
For the main models across the Qwen3.5 hybrid family (2B and 4B), we freeze the pretrained backbone and perform memory-aware post-training on MemOnce-1M for one epoch on an $8\times$ NVIDIA H100 node with Distributed Data Parallel and activation checkpointing, following the configurations in Table~\ref{tab:training_config}. For representation and reranking evaluations, our testing protocol strictly follows EvoEmbedding~\citep{nie2026evoembedding}.

\paragraph{Trainable Modules and Optimization Settings.}
Across the Qwen3.5 hybrid backbones (2B and 4B), post-training adaptation introduces three lightweight trainable components while keeping the base model frozen:
\begin{enumerate}
    \item \textbf{Memory LoRA Adapter:} Low-rank adapters ($r=128$, $\alpha=256$, dropout $0.05$) attached to the attention projections, activated exclusively during memory encoding ($g_\theta$) and query descriptor extraction, and disabled during response generation.
    \item \textbf{Indexing Projection $\phi(\cdot)$:} A two-layer SiLU MLP mapping the terminal token state of each segment from hidden dimension $D$ to $d_e = 1024$ for cosine similarity routing ($\tau = 0.1$, retrieval budget $K = 64$).
    \item \textbf{On-Demand Decoders ($u_\theta$):} For full-attention layers only (e.g., $8$ of $32$ layers in Qwen3.5-4B), trainable copies of the input LayerNorm, $K_{\text{Proj}}^l / V_{\text{Proj}}^l$ projections, and key normalization, initialized from the frozen backbone, decode the active memory state $M_t = [M_t^{\text{long}} \,\|\, G_t^{\text{short}}]$ into layer-specific KV caches $C_t$.
\end{enumerate}
The model is optimized end-to-end under Eq.~\eqref{eq:training_loss} with $\lambda = 0.1$ and $\tau = 0.1$ (computing the contrastive loss over all available historical states when the dialogue fits entirely within $G_t^{\text{short}}$).

\subsection{Memory Benchmarks}
\label{subsec:memory_benchmarks}

We evaluate MemOnce across six complementary benchmarks spanning short-context state updates to million-token multi-session interactions:
\begin{itemize}
    \item \textbf{MemBench}\footnote{\url{https://github.com/import-myself/Membench}}~\citep{tan2025membench}: To assess precise state tracking and fact overwriting without long-context dilution, we adopt the Knowledge Update (KU) subset of MemBench, which comprises $2,000$ evaluation instances with an average history length of approximately $4\text{K}$ tokens.
    \item \textbf{LoCoMo}\footnote{\url{https://huggingface.co/datasets/Percena/locomo-mc10}}~\citep{maharana2024evaluating}: A multi-session conversational memory benchmark consisting of $10$ extended daily-life dialogues, each spanning roughly $30$ sessions. The model assumes the role of one speaker and must concisely answer questions regarding specific biographical and event details of both participants based on the entire conversation history. Following standard evaluation practice, we filter out the unanswerable adversarial category (Category 5), resulting in $1,540$ evaluation questions.
    \item \textbf{LongMemEval}\footnote{\url{https://huggingface.co/datasets/xiaowu0162/longmemeval}}~\citep{wu2025longmemeval}: A comprehensive multi-session memory benchmark covering five core memory capabilities (single-session recall, multi-session synthesis, temporal reasoning, knowledge updates, and abstention), containing $500$ questions each paired with approximately $115\text{K}$ tokens of timestamped multi-session history.
    \item \textbf{PersonaMem-128}~\citep{jiang2025know}: A long-context memory benchmark sharing the same data distribution and four-way multiple-choice task formulation as PersonaMem-1M, designed to evaluate fine-grained recall of user-disclosed facts and preferences over $1,000$ test questions within a $128\text{K}$-token context horizon.
    \item \textbf{PersonaMem-1M}\footnote{\url{https://huggingface.co/datasets/bowen-upenn/PersonaMem-v1}}~\citep{jiang2025know}: An ultra-long-range personalized memory stress test comprising $500$ four-way multiple-choice questions. Each instance requires tracking implicit user facts, evolving preferences, and persona details across $32$ conversational sessions totaling approximately $1\text{M}$ tokens.
    \item \textbf{HotpotQA$^\clubsuit$-1M}: A multi-hop QA benchmark constructed from HotpotQA\footnote{\url{https://huggingface.co/datasets/hotpotqa/hotpot_qa}}~\citep{yang2018hotpotqa} to evaluate long-context memory capabilities. It is synthesized by aggregating $9,848$ raw validation passages (approximately $1.35\text{M}$ tokens) while retaining $500$ open-ended multi-hop questions whose supporting evidence resides completely within the aggregated memory bank.
\end{itemize}

\subsection{Evaluation Protocol}
\label{subsec:judge_prompt}

For multiple-choice benchmarks (\textbf{MemBench}, \textbf{PersonaMem-128}, and \textbf{PersonaMem-1M}), we report exact accuracy. For open-ended QA and conversational benchmarks (\textbf{LoCoMo}, \textbf{LongMemEval}, and \textbf{HotpotQA$^\clubsuit$-1M}), we follow the official LLM-as-a-Judge protocols specified by each benchmark to ensure fair comparison. Specifically, we adopt benchmark-native binary grading prompts (e.g., tailored prompt~\citep{nie2026evoembedding} templates for LongMemEval and LoCoMo) to evaluate whether the generated response entails the ground-truth answer. Empirical verification confirms that employing these official benchmark-aligned prompts yields consistent relative performance trends without affecting the core conclusions. The standard binary evaluation prompt is shown below:

\begin{quote}
\small
\ttfamily
I will give you a question, a correct answer, and a response from a model. Please answer yes if the response contains the correct answer. Otherwise, answer no. If the response is equivalent to the correct answer or contains all the intermediate steps to get the correct answer, you should also answer yes. If the response only contains a subset of the information required by the answer, answer no.\\[6pt]
Question: \{question\}\\[6pt]
Correct Answer: \{answer\}\\[6pt]
Model Response: \{response\}\\[6pt]
Is the model response correct? Answer yes or no only.
\end{quote}

\begin{table*}[!t]
\centering
\caption{\textbf{Ablation study on compression ratio $r$ and long-term linear state persistence across eight downstream benchmarks.} Best results are in \textbf{bold} and second-best are \underline{underlined}.}
\label{tab:ablation_compression}
\resizebox{0.92\textwidth}{!}{
\begin{tabular}{lc|cccccccc|c}
\toprule
\textbf{Linear State} & \textbf{Ratio $r$} & \textbf{PIQA} & \textbf{Hella.} & \textbf{Wino.} & \textbf{ARC-E} & \textbf{ARC-C} & \textbf{SciQ} & \textbf{BoolQ} & \textbf{LAMB.} & \textbf{Avg.} \\
\midrule
\multirow{4}{*}{Persisted} 
 & $1\times$ & 0.669 & \underline{0.419} & \textbf{0.527} & \textbf{0.531} & \underline{0.288} & 0.814 & \underline{0.617} & 0.360 & \textbf{0.5280} \\
 & $2\times$ & 0.673 & 0.416 & 0.493 & \underline{0.525} & 0.278 & \underline{0.845} & 0.600 & \textbf{0.376} & 0.5259 \\
 & $4\times$ & 0.664 & 0.416 & 0.511 & 0.510 & 0.264 & 0.831 & 0.609 & 0.366 & 0.5213 \\
 & $8\times$ & 0.668 & 0.418 & \underline{0.522} & 0.516 & 0.281 & 0.836 & 0.613 & 0.350 & 0.5255 \\
\midrule
\multirow{4}{*}{Discarded} 
 & $1\times$ & \underline{0.676} & \textbf{0.422} & 0.496 & 0.524 & 0.275 & \textbf{0.847} & \textbf{0.619} & 0.352 & \underline{0.5263} \\
 & $2\times$ & 0.668 & 0.414 & 0.518 & 0.514 & \textbf{0.290} & 0.835 & 0.590 & 0.360 & 0.5235 \\
 & $4\times$ & 0.675 & 0.414 & 0.502 & 0.513 & 0.283 & 0.836 & 0.608 & \underline{0.370} & 0.5253 \\
 & $8\times$ & \textbf{0.678} & \underline{0.419} & 0.518 & 0.523 & 0.277 & 0.817 & 0.616 & 0.340 & 0.5235 \\
\bottomrule
\end{tabular}
}
\end{table*}

\section{Ablation Study}
\label{sec:ablation}

\subsection{Compression Ratio}
We first evaluate the trade-off between the compression ratio $r \in \{1, 2, 4, 8\}$ and downstream accuracy in Table~\ref{tab:ablation_compression}. Increasing $r$ from $1\times$ to $8\times$ compresses the persistent memory bank $G_t$ by $8\times$ along the sequence dimension, while causing only a negligible accuracy drop of $0.0025$ ($0.5280 \to 0.5255$ when linear states are persisted) and $0.0028$ ($0.5263 \to 0.5235$ when discarded).
This confirms that the trailing $1/r$ final-layer hidden states effectively encapsulate segment-level semantics, allowing $r=8$ to serve as our default setting.

\subsection{Linear State Persistence}
Table~\ref{tab:ablation_compression} also examines whether the recurrent states of linear layers need to be maintained across historical segments. The \textit{Persisted} setting carries linear recurrent states across segments alongside $G_t$, whereas the \textit{Discarded} setting resets linear states at segment boundaries and relies solely on the decoded KV caches from $M_t$ for long-range context. Across all compression ratios, both settings yield nearly identical performance (e.g., $0.5255$ vs.\ $0.5235$ at $r=8$), demonstrating that long-range information is predominantly preserved by our expandable memory bank $G_t$ rather than fixed-capacity linear states. We therefore discard cross-segment linear states by default, which eliminates multi-layer recurrent state storage.

\subsection{Top-$K$ Retrieval Budget}
We next analyze the sensitivity of MemOnce to the retrieval budget $K \in \{0, 16, 32, 64, 128\}$ across five long-context memory benchmarks (Table~\ref{tab:ablation_topk}). At $K=0$, the model relies solely on the short-term window $G_t^{\text{short}}$ without long-term retrieval, resulting in a low average score of $36.83$. Retrieving just $K=16$ long-term representations yields a substantial gain of $+22.89$ points ($59.72$), verifying the high information density of our compact memory states. Performance peaks at $K=64$ (\textbf{60.07}), capturing sufficient multi-session evidence for multi-hop and temporal queries in LoCoMo ($67.01$) and LongMemEval ($47.40$). Further expanding $K$ to $128$ slightly reduces the average score to $58.86$ while doubling the retrieval footprint, as excessive low-relevance states introduce contextual noise. Consequently, we adopt $K=64$ as the default retrieval budget.

\begin{table*}[!t]
\centering
\caption{\textbf{Ablation study on the retrieval budget $K$ across long-context memory benchmarks.} Best results are in \textbf{bold} and second-best are \underline{underlined}.}
\label{tab:ablation_topk}
\resizebox{0.85\textwidth}{!}{
\begin{tabular}{lc|ccccc|c}
\toprule
\textbf{Model} & \textbf{Top-$K$} & \textbf{MemBench} & \textbf{LoCoMo} & \textbf{LongMemEval} & \textbf{PersonaMem-128} & \textbf{PersonaMem-1M} & \textbf{Mem. Avg.} \\
\midrule
MemOnce-4B & $K=0$   & 72.49 & 20.06 & 14.00 & 40.20 & 37.40 & 36.83 \\
MemOnce-4B & $K=16$  & \textbf{84.94} & 65.06 & \underline{47.00} & \textbf{53.60} & \underline{48.00} & \underline{59.72} \\
MemOnce-4B & $K=32$  & \underline{84.64} & \underline{66.23} & 45.20 & \textbf{53.60} & \textbf{48.80} & 59.69 \\
\rowcolor{rowdeep} \textbf{MemOnce-4B} & \textbf{$K=64$} & 84.34 & \textbf{67.01} & \textbf{47.40} & \textbf{53.60} & \underline{48.00} & \textbf{60.07} \\
MemOnce-4B & $K=128$ & \textbf{84.94} & 65.97 & 45.60 & \underline{51.00} & 46.80 & 58.86 \\
\bottomrule
\end{tabular}
}
\end{table*}

\begin{figure*}[!t]
\centering
\includegraphics[width=\textwidth]{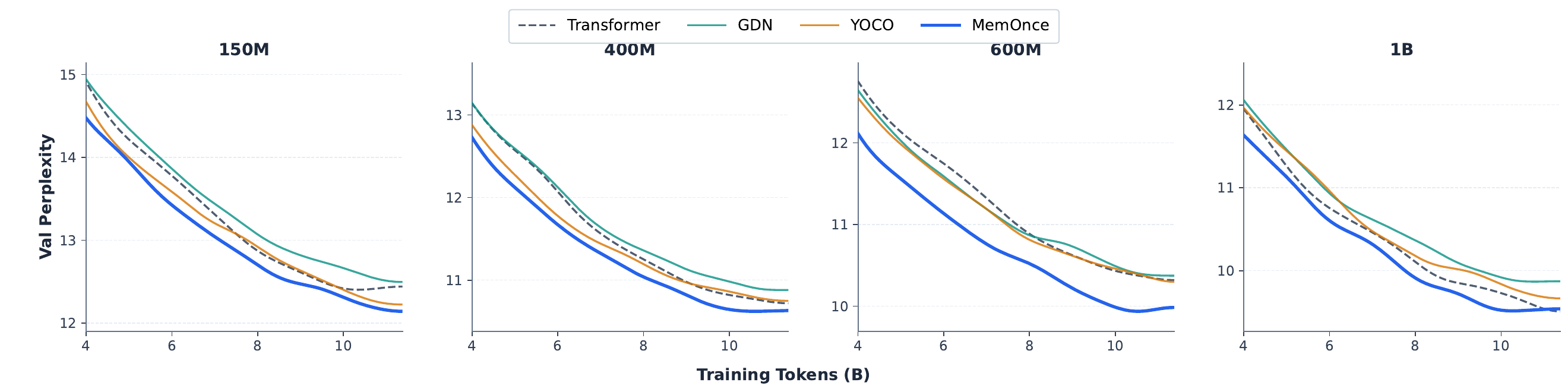}
\caption{\textbf{Pre-training dynamics across the four parameter tiers.} Gaussian-smoothed validation perplexity on FineWeb-Edu-10B at the $\sim$150M, $\sim$400M, $\sim$600M, and $\sim$1B scales. MemOnce descends fastest and reaches the lowest validation perplexity at 150M--600M and stays on par with the full-attention Transformer during training at 1B, where it attains the lower final evaluated perplexity (Table~\ref{tab:scaling_results}).}
\label{fig:scaling_curves}
\end{figure*}

\subsection{Retrieval Efficiency}
Finally, Table~\ref{tab:ablation_efficiency} compares the query-time retrieval latency breakdown and PCIe data transfer of MemOnce against RAG and M+ at a 128K context ($256$ segments of $512$ tokens). MemOnce encodes the query via a single prompt forward pass through the backbone ($60.39\text{ ms}$, with the short-term state injected as decoded KV) and loads the top-$64$ compact memory representations ($20.00\text{ MB}$) from SSD in $3.85\text{ ms}$, totaling {64.24 ms} ($63.0\text{ ms}$ with OS page cache and $60.7\text{ ms}$ when $G_t^{\text{long}}$ resides on GPU). In contrast, M+ performs layer-wise retrieval across all $32$ layers during the forward pass, transferring $1,150.00\text{ MB}$ of multi-layer states from CPU to GPU and incurring $772.50\text{ ms}$ latency. Meanwhile, RAG-Rewrite~\citep{ma2023query} first invokes an LLM to rewrite the conversational query with historical context ($1081.84\text{ ms}$), before searching and reading the retrieved text, totaling $1094.30\text{ ms}$, alongside additional prefill compute over retrieved tokens. Overall, MemOnce achieves a $17.0\times$ latency speedup over RAG-Rewrite and a $12.0\times$ speedup over M+.

\begin{table}[!t]
\centering
\caption{\textbf{Query-time retrieval overhead of RAG-Rewrite, M+, and MemOnce at a 128K-token context.} Latency is broken down by pipeline stage. Off-pool reads use SSD cold-read latency. M+ retrieves layer-wise inside the forward pass and has no separate encoding or reading stage.}
\label{tab:ablation_efficiency}
\resizebox{.8\columnwidth}{!}{
\begin{tabular}{l|ccc|c|c}
\toprule
\textbf{System} & \textbf{Encode} & \textbf{Search \& Load} & \textbf{Read} & \textbf{Total Latency} & \textbf{Data Transfer} \\
\midrule
RAG-Rewrite~\citep{ma2023query} & 1081.84 ms & 0.46 ms & 12.00 ms & 1094.30 ms & 0.13 MB \\
M+~\citep{wang2025m} & --- & 772.50 ms & --- & 772.50 ms & 1,150.00 MB \\
\rowcolor{rowdeep} \textbf{MemOnce (Ours)} & \textbf{60.39 ms} & \textbf{3.85 ms} & \textbf{---} & \textbf{64.24 ms} & \textbf{20.00 MB} \\
\bottomrule
\end{tabular}
}
\end{table}

\section{Additional Experimental Results and Analysis}
\label{app:results}

\subsection{Scaling Behavior}
To verify that MemOnce scales effectively across model capacities, we evaluate four parameter tiers ($\sim$150M, $\sim$400M, $\sim$600M, and $\sim$1B) pre-trained from scratch on FineWeb-Edu-10B against a standard full-attention Transformer, pure Gated DeltaNet (GDN)~\citep{yang2025gated}, and YOCO~\citep{sun2024you} in Table~\ref{tab:scaling_results}, complementing the comprehensive 400M architectural comparison in Table~\ref{tab:full_400m_results}. At the 150M scale, MemOnce achieves the lowest perplexity on four of the five corpora, reducing WikiText perplexity from $25.64$ (Transformer) and $31.22$ (GDN) to $23.40$, while maintaining competitive downstream accuracy.
As capacity scales to 400M, 600M, and 1B parameters, both uncompressed ($r=1$) and $8\times$ compressed ($r=8$) variants of MemOnce consistently achieve the top-two average accuracies (\textbf{Avg.}) across all eight zero-shot benchmarks while maintaining strong language modeling perplexity.
Specifically, whereas pure GDN suffers from severe long-range perplexity degradation on WikiText and Penn Treebank (PTB) due to fixed-state bottlenecks, MemOnce scales smoothly to achieve the best validation perplexity ($10.16$ at 603M and $9.65$ at 945M) and highest overall downstream accuracy ($0.5409$ and $0.5515$). Crucially, the gap between $r=8$ and $r=1$ narrows with scale, enabling the $8\times$ compressed model to likewise outperform the 1B full-attention Transformer with fewer parameters. Figure~\ref{fig:scaling_curves} traces the full pre-training dynamics at the four scales.

\begin{table*}[!t]
\centering
\caption{\textbf{Pre-training scaling comparison from $\sim$150M to $\sim$1B parameters.} We report validation perplexity ($\downarrow$) and zero-shot accuracy ($\uparrow$) across eight downstream benchmarks. MemOnce is evaluated under both the uncompressed configuration ($r=1$) and the $8\times$ compressed configuration ($r=8$) across all scales. Best results within each scale are in \textbf{bold} and second-best are \underline{underlined}.}
\label{tab:scaling_results}
\resizebox{\textwidth}{!}{
\begin{tabular}{ll|ccccc|cccccccc|c}
\toprule
\textbf{Model} & \textbf{Params} & \textbf{Val PPL $\downarrow$} & \textbf{WT2 $\downarrow$} & \textbf{WikiText $\downarrow$} & \textbf{LAMB. $\downarrow$} & \textbf{PTB $\downarrow$} & \textbf{PIQA} & \textbf{Hella.} & \textbf{Wino.} & \textbf{ARC-E} & \textbf{ARC-C} & \textbf{SciQ} & \textbf{BoolQ} & \textbf{LAMB.} & \textbf{Avg. $\uparrow$} \\
\midrule
\multicolumn{16}{l}{\textit{\textbf{$\sim$150M Scale}}} \\
Transformer & 146M & 12.65 & 16.55 & \underline{25.64} & \underline{36.49} & 171.91 & 0.642 & \underline{0.362} & \underline{0.519} & \textbf{0.475} & \underline{0.253} & \textbf{0.799} & 0.579 & 0.294 & \textbf{0.4904} \\
GDN~\citep{yang2025gated} & 174M & 12.82 & 17.26 & 31.22 & 37.40 & 246.51 & 0.642 & 0.355 & 0.516 & 0.470 & \textbf{0.259} & 0.786 & 0.569 & 0.283 & 0.4850 \\
YOCO~\citep{sun2024you} & 152M & \underline{12.63} & \underline{16.39} & 29.35 & 36.98 & 225.30 & 0.632 & 0.330 & \textbf{0.530} & 0.460 & 0.249 & 0.787 & \textbf{0.600} & 0.250 & 0.4798 \\
\rowcolor{rowdeep} \textbf{MemOnce ($r=8$)} & 158M & 12.64 & 16.77 & 25.81 & \textbf{36.17} & \textbf{166.99} & \textbf{0.648} & 0.355 & 0.514 & 0.469 & 0.250 & \underline{0.791} & \underline{0.581} & \underline{0.306} & \underline{0.4893} \\
\rowcolor{rowdeep} \textbf{MemOnce ($r=1$)} & 158M & \textbf{12.46} & \textbf{16.14} & \textbf{23.40} & 36.55 & \underline{171.67} & \underline{0.646} & \textbf{0.364} & 0.506 & \underline{0.474} & \underline{0.253} & \underline{0.791} & 0.548 & \textbf{0.311} & 0.4866 \\
\midrule
\multicolumn{16}{l}{\textit{\textbf{$\sim$400M Scale}}} \\
Transformer & 365M & 10.97 & 13.68 & \underline{20.90} & 33.24 & \textbf{134.34} & 0.657 & 0.404 & \underline{0.515} & \underline{0.524} & \textbf{0.282} & 0.832 & 0.582 & \underline{0.352} & 0.5185 \\
GDN~\citep{yang2025gated} & 410M & 11.11 & 14.24 & 26.03 & 33.37 & 223.14 & \underline{0.676} & 0.406 & 0.503 & \textbf{0.541} & 0.272 & \underline{0.842} & 0.606 & 0.340 & 0.5231 \\
YOCO~\citep{sun2024you} & 387M & 10.97 & \underline{13.63} & 24.38 & 35.18 & 346.33 & 0.674 & 0.407 & 0.511 & 0.516 & 0.265 & 0.826 & 0.597 & \textbf{0.361} & 0.5197 \\
\rowcolor{rowdeep} \textbf{MemOnce ($r=8$)} & 400M & \underline{10.89} & 14.05 & 21.08 & \textbf{32.45} & 151.80 & \textbf{0.678} & \underline{0.419} & \textbf{0.518} & 0.523 & \underline{0.277} & 0.817 & \underline{0.616} & 0.340 & \underline{0.5235} \\
\rowcolor{rowdeep} \textbf{MemOnce ($r=1$)} & 400M & \textbf{10.77} & \textbf{13.33} & \textbf{19.28} & \underline{32.69} & \underline{144.83} & \underline{0.676} & \textbf{0.422} & 0.496 & \underline{0.524} & 0.275 & \textbf{0.847} & \textbf{0.619} & \underline{0.352} & \textbf{0.5263} \\
\midrule
\multicolumn{16}{l}{\textit{\textbf{$\sim$600M Scale}}} \\
Transformer & 516M & 10.57 & 13.16 & 19.83 & 31.55 & \textbf{127.64} & 0.661 & 0.422 & 0.509 & 0.505 & 0.287 & 0.832 & 0.542 & \underline{0.359} & 0.5146 \\
GDN~\citep{yang2025gated} & 587M & 10.61 & 13.44 & 24.77 & 32.16 & 221.83 & 0.667 & 0.422 & 0.515 & 0.516 & 0.275 & 0.827 & \textbf{0.610} & 0.358 & 0.5237 \\
YOCO~\citep{sun2024you} & 580M & 10.52 & 13.01 & 20.97 & 32.14 & 145.74 & \underline{0.673} & 0.424 & 0.509 & \underline{0.548} & \underline{0.300} & 0.841 & 0.574 & 0.330 & 0.5249 \\
\rowcolor{rowdeep} \textbf{MemOnce ($r=8$)} & 603M & \underline{10.34} & \underline{12.87} & \underline{19.52} & \underline{31.04} & 144.66 & \textbf{0.680} & \underline{0.437} & \underline{0.527} & 0.546 & 0.299 & \textbf{0.848} & \underline{0.597} & 0.355 & \underline{0.5361} \\
\rowcolor{rowdeep} \textbf{MemOnce ($r=1$)} & 603M & \textbf{10.16} & \textbf{12.35} & \textbf{18.08} & \textbf{30.70} & \underline{139.43} & \underline{0.673} & \textbf{0.440} & \textbf{0.530} & \textbf{0.551} & \textbf{0.301} & \underline{0.845} & 0.591 & \textbf{0.396} & \textbf{0.5409} \\
\midrule
\multicolumn{16}{l}{\textit{\textbf{$\sim$1B Scale}}} \\
Transformer & 1007M & \underline{9.78} & \underline{11.70} & \underline{17.14} & 29.98 & \textbf{113.45} & \underline{0.687} & 0.456 & 0.526 & 0.552 & \textbf{0.312} & 0.858 & \textbf{0.608} & 0.409 & 0.5512 \\
GDN~\citep{yang2025gated} & 971M & 9.98 & 12.35 & 23.11 & 30.77 & 211.12 & 0.685 & 0.452 & 0.523 & \underline{0.556} & 0.305 & 0.845 & 0.545 & 0.371 & 0.5350 \\
YOCO~\citep{sun2024you} & 977M & 9.87 & 11.88 & 18.42 & 30.47 & 146.47 & 0.681 & 0.454 & 0.522 & 0.550 & \underline{0.306} & 0.855 & \underline{0.596} & 0.407 & 0.5462 \\
\rowcolor{rowdeep} \textbf{MemOnce ($r=8$)} & 945M & 9.83 & 12.12 & 18.11 & \textbf{29.59} & 143.98 & \textbf{0.688} & \underline{0.461} & \textbf{0.537} & \textbf{0.564} & 0.298 & \underline{0.875} & 0.569 & \textbf{0.420} & \underline{0.5513} \\
\rowcolor{rowdeep} \textbf{MemOnce ($r=1$)} & 945M & \textbf{9.65} & \textbf{11.52} & \textbf{17.08} & \underline{29.93} & \underline{128.41} & 0.678 & \textbf{0.466} & \underline{0.530} & 0.546 & 0.300 & \textbf{0.888} & 0.586 & \underline{0.418} & \textbf{0.5515} \\
\bottomrule
\end{tabular}
}
\end{table*}

\subsection{Comparison with RAG}
Table~\ref{tab:memory_rag} compares MemOnce with a RAG baseline that retrieves the top-8 most relevant historical dialogues using All-MiniLM-L6-v2. MemOnce achieves better overall performance ($60.07$ vs. $52.87$).
While RAG achieves a higher score on LongMemEval by directly prepending uncompressed raw text chunks for verbatim single-session extraction, MemOnce outperforms RAG on four out of five benchmarks and improves the overall average by $+7.20$ points under an $8\times$ sequence compression ratio.

\begin{table*}[!t]
\centering
\caption{\textbf{Comparison with RAG on five memory benchmarks.} Best results are in \textbf{bold} and second-best are \underline{underlined}.}
\label{tab:memory_rag}
\resizebox{.8\textwidth}{!}{
\begin{tabular}{lc|ccccc|c}
\toprule
\multirow{2}{*}{\textbf{Model}} & \multirow{2}{*}{\textbf{Size}} & \textbf{MemBench} & \textbf{LoCoMo} & \textbf{LongMemEval} & \textbf{PersonaMem-128} & \textbf{PersonaMem-1M} & \textbf{Mem. Avg.} \\
& & ($\sim$4K) $\uparrow$ & ($\sim$21K) $\uparrow$ & ($\sim$115K) $\uparrow$ & ($\sim$128K) $\uparrow$ & ($\sim$1M) $\uparrow$ & $\uparrow$ \\
\midrule
Qwen3.5-4B (Full Context) & 4B & \underline{82.64} & \textbf{69.55} & 42.20 & 43.50 & 33.60 & \underline{54.30} \\
RAG~\citep{lewis2020rag} & 4B & 70.69 & 45.06 & \textbf{52.40} & \underline{51.00} & \underline{45.20} & 52.87 \\
\rowcolor{rowdeep} \textbf{MemOnce (Ours)} & 4B & \textbf{84.34} & \underline{67.01} & \underline{47.40} & \textbf{53.60} & \textbf{48.00} & \textbf{60.07} \\
\bottomrule
\end{tabular}
}
\end{table*}

\subsection{Long-Context Memory}
\label{subsec:hotpot_1m}

To further validate the memory capabilities of MemOnce at the $1\text{M}$-token scale beyond multiple-choice tasks, we evaluate on \textbf{HotpotQA$^\clubsuit$-1M}, which concatenates $9,848$ context passages across validation samples ($\sim 1.35\text{M}$ tokens) and evaluates on $500$ open-ended multi-hop questions with complete in-bank supporting evidence. To verify that the improvements stem from genuine memory utilization rather than parametric overfitting, we report both zero-context accuracy (\textbf{No Mem.}) and the resulting net memory gain ($\Delta$). As shown in Table~\ref{tab:hotpot_1m}, the post-trained MemOnce-2B and MemOnce-4B models achieve the highest memory gains ($\Delta = +41.00$ and $+39.80$). Moreover, MemOnce-4B surpasses RAG-4B ($59.40$ vs.\ $52.20$) and outperforms existing memory systems by a substantial margin, further demonstrating the effectiveness of MemOnce.

\begin{table}[!t]
\centering
\caption{\textbf{Memory capability evaluation on HotpotQA$^\clubsuit$-1M.} We report accuracy (\%) with memory (\textbf{With Mem.}), without memory (\textbf{No Mem.}), and net memory gain ($\mathbf{\Delta}$). Best results are in \textbf{bold} and second-best are \underline{underlined}.}
\label{tab:hotpot_1m}
\resizebox{0.75\columnwidth}{!}{
\begin{tabular}{lc|ccc}
\toprule
\textbf{Model} & \textbf{Size} & \textbf{With Mem. ($\text{ctx}$) $\uparrow$} & \textbf{No Mem. ($\text{noctx}$)} & \textbf{Gain ($\Delta$) $\uparrow$} \\
\midrule
\multicolumn{5}{l}{\textit{\textbf{Full Context \& Retrieval Baselines}}} \\
Qwen3.5-2B & 2B & 23.80 & 8.60 & +15.20 \\
Qwen3.5-4B & 4B & 26.60 & 10.60 & +16.00 \\
RAG~\citep{lewis2020rag} & 4B & \underline{52.20} & 15.60 & +36.60 \\
\midrule
\multicolumn{5}{l}{\textit{\textbf{Native Memory Models}}} \\
$\delta$-Mem~\citep{lei2026delta} & 4B & 21.60 & 10.60 & +11.00 \\
EvoEmbedding~\citep{nie2026evoembedding} & 4B & 21.40 & 11.00 & +10.40 \\
Metis~\citep{zhang2026metis} & 4B & 18.20 & 10.60 & +7.60 \\
M+~\citep{wang2025m} & 8B & 33.40 & 7.20 & +26.20 \\
\cmidrule{1-5}
$\delta$-Mem$^\dagger$~\citep{lei2026delta} & 4B & 26.20 & 18.80 & +7.40 \\
EvoEmbedding$^\dagger$~\citep{nie2026evoembedding} & 4B & 28.20 & 20.00 & +8.20 \\
Metis$^\dagger$~\citep{zhang2026metis} & 4B & 26.60 & 16.40 & +10.20 \\
\midrule
\rowcolor{rowdeep} \textbf{MemOnce (Ours)} & 2B & 49.80 & 8.80 & \textbf{+41.00} \\
\rowcolor{rowdeep} \textbf{MemOnce (Ours)} & 4B & \textbf{59.40} & 19.60 & \underline{+39.80} \\
\bottomrule
\end{tabular}
}
\end{table}

\subsection{Inference Efficiency Across Context Lengths}
Table~\ref{tab:efficiency_scaling} benchmarks prefill latency, computational FLOPs, decoding throughput, and active versus offloaded memory footprints from $32\text{K}$ to $1\text{M}$ tokens. Crucially, these performance disparities stem from fundamental architectural designs rather than engineering or implementation artifacts.
The full-context Qwen3.5 baseline retains quadratic attention, rendering its prefill competitive only at short lengths and causing decoding throughput to collapse once the KV cache exceeds the memory-bandwidth budget.
Meanwhile, the official implementation of $\delta$-Mem constructs its associative states through full-context forward passes, negating its theoretical linear-time advantage during inference.
Metis is likewise linear-time and matches MemOnce in decoding speed, but its state updates proceed on overly large blocks that underutilize GPU parallelism, incurring a roughly $3\times$ prefill overhead.
EvoEmbedding maintains separate generation and retrieval adapters, requiring two context forward passes when switching between them.
Furthermore, M+ performs layer-wise retrieval during the forward pass, triggering frequent CPU--GPU transfers that throttle compute utilization to ${\sim}10$ TFLOP/s.
In contrast, MemOnce bounds chunk attention to $4\text{K}$-token windows at $167$--$180$ TFLOP/s utilization and holds its active memory state to at most $C = 5120$ tokens while offloading the growing pool to SSD, thereby sustaining linear prefill and invariant decoding throughput across the entire $1\text{M}$ horizon.

\begin{table*}[!t]
\centering
\caption{\textbf{Inference efficiency scaling from 32K to 1M tokens (4B scale).} We report prefill time (s), estimated FLOPs (with effective hardware throughput in TFLOP/s), decoding throughput (tokens/s), active memory cache footprint, and offloaded SSD/CPU memory pool size. Single-GPU (80GB H100) out-of-memory (OOM) baselines are evaluated with multi-GPU.}
\label{tab:efficiency_scaling}
\resizebox{0.92\textwidth}{!}{
\begin{tabular}{l|ccccc}
\toprule
\textbf{Method} & \textbf{Prefill Time (s) $\downarrow$} & \textbf{Prefill FLOPs (Throughput)} & \textbf{Decode (tok/s) $\uparrow$} & \textbf{Active Memory Cache $\downarrow$} & \textbf{Offloaded SSD/CPU} \\
\midrule
\multicolumn{6}{l}{\textit{\textbf{32K Context Length}}} \\
Qwen3.5 (Baseline) & \textbf{0.87} & 308T (354 TF/s) & 26.5 & 1.05 GB & --- \\
M+~\citep{wang2025m} & 24.40 & 466T (19 TF/s) & 24.8 & 2.50 GB & 4.60 GB \\
$\delta$-Mem~\citep{lei2026delta} & 2.24 & 634T (283 TF/s) & 21.6 & 4.5 KB & --- \\
EvoEmbedding~\citep{nie2026evoembedding} & 5.02 & 525T (105 TF/s) & \underline{28.2} & 2.6 MB & --- \\
Metis~\citep{zhang2026metis} & 4.22 & 221T (52 TF/s) & 27.5 & 80.1 MB & --- \\
\rowcolor{rowdeep} \textbf{MemOnce (Ours)} & \underline{1.38} & 230T (167 TF/s) & \textbf{29.4} & 20.0 MB & 20.0 MB \\
\midrule
\multicolumn{6}{l}{\textit{\textbf{128K Context Length}}} \\
Qwen3.5 (Baseline) & \underline{6.66} & 2.29P (344 TF/s) & 18.7 & 4.05 GB & --- \\
M+~\citep{wang2025m} & 105.42 & 1.86P (18 TF/s) & 24.4 & 2.50 GB & 16.60 GB \\
$\delta$-Mem~\citep{lei2026delta} & 19.91 & 7.28P (366 TF/s) & 21.2 & 4.5 KB & --- \\
EvoEmbedding~\citep{nie2026evoembedding} & 20.04 & 2.10P (105 TF/s) & \textbf{29.0} & 2.6 MB & --- \\
Metis~\citep{zhang2026metis} & 16.40 & 884T (54 TF/s) & 28.0 & 80.1 MB & --- \\
\rowcolor{rowdeep} \textbf{MemOnce (Ours)} & \textbf{5.11} & 921T (180 TF/s) & \textbf{29.0} & 25.0 MB & 80.0 MB \\
\midrule
\multicolumn{6}{l}{\textit{\textbf{512K Context Length}}} \\
Qwen3.5 (Baseline) & 77.47 & 26.03P (336 TF/s) & 12.8 & 16.05 GB & --- \\
M+~\citep{wang2025m} & 645.66 & 7.46P (12 TF/s) & 23.6 & 2.50 GB & 35.90 GB \\
$\delta$-Mem~\citep{lei2026delta} & 304.26 & 105.14P (346 TF/s) & 16.5 & 4.5 KB & --- \\
EvoEmbedding~\citep{nie2026evoembedding} & 81.26 & 8.40P (103 TF/s) & \underline{28.7} & 2.6 MB & --- \\
Metis~\citep{zhang2026metis} & \underline{65.82} & 3.53P (54 TF/s) & 28.3 & 80.1 MB & --- \\
\rowcolor{rowdeep} \textbf{MemOnce (Ours)} & \textbf{20.98} & 3.69P (176 TF/s) & \textbf{29.3} & 25.0 MB & 320.0 MB \\
\midrule
\multicolumn{6}{l}{\textit{\textbf{1M Context Length}}} \\
Qwen3.5 (Baseline) & 286.86 & 97.10P (338 TF/s) & 8.4 & 32.05 GB & --- \\
M+~\citep{wang2025m} & 1479.50 & 14.92P (10 TF/s) & 21.5 & 2.50 GB & 35.90 GB \\
$\delta$-Mem~\citep{lei2026delta} & 1065.88 & 412.94P (387 TF/s) & 18.0 & 4.5 KB & --- \\
EvoEmbedding~\citep{nie2026evoembedding} & 160.33 & 16.81P (105 TF/s) & \underline{29.1} & 2.6 MB & --- \\
Metis~\citep{zhang2026metis} & \underline{132.26} & 7.07P (53 TF/s) & 28.0 & 80.1 MB & --- \\
\rowcolor{rowdeep} \textbf{MemOnce (Ours)} & \textbf{41.36} & 7.37P (178 TF/s) & \textbf{29.3} & 25.0 MB & 640.0 MB \\
\bottomrule
\end{tabular}
}
\end{table*}

\subsection{Architectural Comparison}
Table~\ref{tab:full_400m_results} presents the complete pre-training evaluation on FineWeb-Edu-10B, comparing MemOnce against standard full-attention Transformers (365M and 516M), Sliding-Window Attention (SWA), cross-layer KV sharing (YOCO~\citep{sun2024you} and LastKV~\citep{cai2026last}), latent KV compression (MLA~\citep{liu2024deepseek}), recurrent linear models (DeltaNet, Gated DeltaNet~\citep{yang2025gated}, and Kimi-Linear~\citep{team2026kimi}), segment-level memory caching (MemCache-GRM~\citep{behrouz2026memory}), and a standard hybrid backbone. Two key findings emerge:
(1) \textbf{Superiority over canonical attention and recurrent baselines:} All MemOnce variants ($r \in \{1, 2, 4, 8\}$) consistently outperform the 365M Transformer ($0.5185$ Avg.), SWA ($0.5160$), MLA ($0.5187$), YOCO ($0.5197$), and even the $1.4\times$ larger 516M Transformer ($0.5146$) in overall downstream accuracy, while avoiding the severe long-range perplexity spikes of pure recurrent architectures on WikiText and PTB ($26.03 / 223.14$ for GDN and $26.70 / 223.33$ for Kimi-Linear vs.\ $19.28\text{--}21.08 / 138.31\text{--}151.80$ for MemOnce). Meanwhile, directly reusing raw final-layer KV states in LastKV yields $11.13$ validation perplexity and $0.5109$ Avg., trailing all MemOnce variants, while MemCache-GRM degrades to $15.60$ validation perplexity and $0.4578$ Avg.
(2) \textbf{Resilience across compression ratios:} MemOnce maintains remarkably stable perplexity and downstream performance from $r=1$ up to $r=8$, reaching \textbf{0.5263} Avg.\ at $r=1$ (the best across all models).

\begin{table*}[!t]
\centering
\caption{\textbf{Comprehensive pre-training comparison of architectural baselines and MemOnce.} We report language modeling perplexity ($\downarrow$) across five corpora and zero-shot accuracy ($\uparrow$) across eight downstream benchmarks (\textbf{Avg.}).}
\label{tab:full_400m_results}
\resizebox{\textwidth}{!}{
\begin{tabular}{ll|ccccc|cccccccc|c}
\toprule
\textbf{Model} & \textbf{Params} & \textbf{Val PPL $\downarrow$} & \textbf{WT2 $\downarrow$} & \textbf{WikiText $\downarrow$} & \textbf{LAMB. $\downarrow$} & \textbf{PTB $\downarrow$} & \textbf{PIQA} & \textbf{Hella.} & \textbf{Wino.} & \textbf{ARC-E} & \textbf{ARC-C} & \textbf{SciQ} & \textbf{BoolQ} & \textbf{LAMB.} & \textbf{Avg. $\uparrow$} \\
\midrule
\multicolumn{16}{l}{\textit{\textbf{Transformers \& Attention Variants}}} \\
Transformer & 516M & \textbf{10.57} & \underline{13.16} & 19.83 & \textbf{31.55} & \textbf{127.64} & 0.661 & \textbf{0.422} & 0.509 & 0.505 & \underline{0.287} & 0.832 & 0.542 & 0.359 & 0.5146 \\
Transformer & 365M & 10.97 & 13.68 & 20.90 & 33.24 & 134.34 & 0.657 & 0.404 & 0.515 & 0.524 & 0.282 & 0.832 & 0.582 & 0.352 & 0.5185 \\
SWA & 365M & 11.00 & 13.72 & 20.52 & 32.69 & \underline{131.13} & 0.656 & 0.410 & 0.514 & 0.501 & 0.274 & 0.832 & 0.597 & 0.345 & 0.5160 \\
MLA~\citep{liu2024deepseek} & 387M & 11.20 & 13.93 & 23.07 & 33.37 & 135.89 & 0.670 & 0.397 & 0.507 & 0.506 & 0.277 & 0.832 & \textbf{0.619} & 0.342 & 0.5187 \\
\midrule
\multicolumn{16}{l}{\textit{\textbf{Linear \& Recurrent Architectures}}} \\
DeltaNet & 309M & 11.69 & 15.07 & 23.75 & 33.71 & 151.01 & 0.656 & 0.379 & 0.510 & 0.491 & 0.268 & 0.807 & 0.603 & 0.308 & 0.5028 \\
GDN~\citep{yang2025gated} & 410M & 11.11 & 14.24 & 26.03 & 33.37 & 223.14 & \underline{0.676} & 0.406 & 0.503 & \textbf{0.541} & 0.272 & \underline{0.842} & 0.606 & 0.340 & 0.5231 \\
Kimi-Linear~\citep{team2026kimi} & 315M & 11.15 & 14.22 & 26.70 & 33.44 & 223.33 & 0.664 & 0.402 & \textbf{0.521} & \underline{0.531} & 0.284 & 0.840 & 0.602 & 0.357 & 0.5250 \\
\midrule
\multicolumn{16}{l}{\textit{\textbf{Efficient \& Memory Architectures}}} \\
Hybrid Backbone & 394M & \underline{10.70} & \textbf{13.04} & \underline{19.56} & 32.61 & 146.76 & 0.665 & \underline{0.420} & 0.517 & 0.514 & 0.271 & 0.834 & 0.587 & 0.351 & 0.5199 \\
YOCO~\citep{sun2024you} & 387M & 10.97 & 13.63 & 24.38 & 35.18 & 346.33 & 0.674 & 0.407 & 0.511 & 0.516 & 0.265 & 0.826 & 0.597 & \underline{0.361} & 0.5197 \\
LastKV~\citep{cai2026last} & 365M & 11.13 & 13.78 & 21.12 & 34.04 & 148.66 & 0.655 & 0.400 & 0.505 & 0.509 & 0.278 & 0.815 & 0.588 & 0.337 & 0.5109 \\
MemCache-GRM~\citep{behrouz2026memory} & 309M & 15.60 & 22.77 & 29.84 & 42.47 & 178.42 & 0.626 & 0.335 & 0.499 & 0.475 & 0.257 & 0.728 & 0.566 & 0.176 & 0.4578 \\
\midrule
\multicolumn{16}{l}{\textit{\textbf{MemOnce (Ours) at Different Compression Ratios}}} \\
\rowcolor{rowdeep} MemOnce ($r=1$) & 400M & 10.77 & 13.33 & \textbf{19.28} & 32.69 & 144.83 & \underline{0.676} & \textbf{0.422} & 0.496 & 0.524 & 0.275 & \textbf{0.847} & \textbf{0.619} & 0.352 & \textbf{0.5263} \\
\rowcolor{rowdeep} MemOnce ($r=2$) & 400M & 10.85 & 13.69 & 19.84 & \underline{32.03} & 138.31 & 0.668 & 0.414 & \underline{0.518} & 0.514 & \textbf{0.290} & 0.835 & 0.590 & 0.360 & 0.5235 \\
\rowcolor{rowdeep} MemOnce ($r=4$) & 400M & 10.88 & 13.97 & 20.66 & 32.29 & 147.16 & 0.675 & 0.414 & 0.502 & 0.513 & 0.283 & 0.836 & 0.608 & \textbf{0.370} & \underline{0.5253} \\
\rowcolor{rowdeep} MemOnce ($r=8$) & 400M & 10.89 & 14.05 & 21.08 & 32.45 & 151.80 & \textbf{0.678} & 0.419 & \underline{0.518} & 0.523 & 0.277 & 0.817 & \underline{0.616} & 0.340 & 0.5235 \\
\bottomrule
\end{tabular}
}
\end{table*}

\subsection{Qualitative Analysis}
\begin{figure*}[!t]
\centering
\includegraphics[width=.8\textwidth]{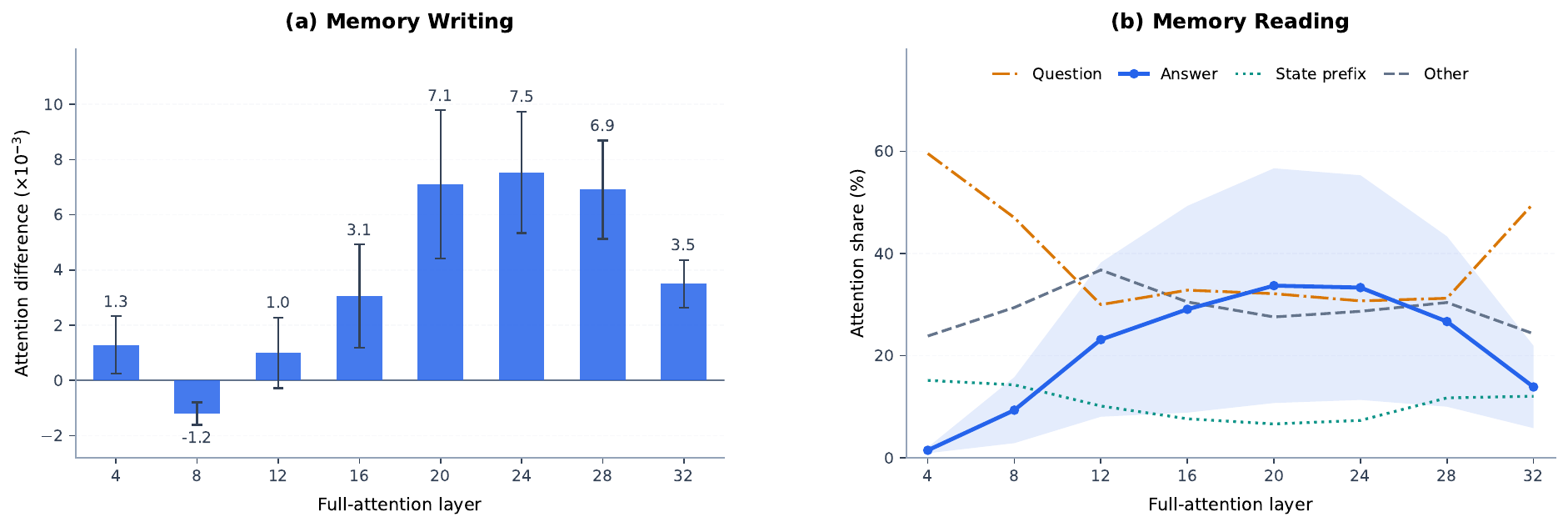}
\caption{\textbf{Attention dynamics during memory writing and reading across full-attention layers.} \textbf{(a) Memory Writing:} Attention difference from trailing memory tokens to factual spans versus filler spans. Middle-to-deep layers selectively attend to salient facts over background noise. \textbf{(b) Memory Reading:} Attention distribution (\%) from generated answer tokens across \textit{Question}, \textit{Answer}, remaining retrieved blocks (\textit{Other}), and the short-term window (\textit{State prefix}). Attention to the answer-carrying block peaks at layers 20--24. Error bars and shading indicate $\pm 1$ standard deviation.}
\label{fig:memory_attention_analysis}
\end{figure*}

We extract attention weights across the eight full-attention layers (layers $4, 8, \dots, 32$) of MemOnce-4B to inspect both memory pathways. For \textit{memory writing}, we feed $512$-token segments embedding factual statements amidst filler text through the memory encoder and measure the attention difference from the trailing $64$ memory tokens to the factual span versus a length-matched filler span. For \textit{memory reading} on LoCoMo, we construct the context from the $16$-block short-term window (\textit{State prefix}), the top-$16$ retrieved long-term blocks, and the query (\textit{Question}), and partition the attention of teacher-forced answer tokens across the answer-carrying block (\textit{Answer}), the remaining $15$ retrieved blocks (\textit{Other}), \textit{Question}, and \textit{State prefix}. Writing results are averaged over $60$ synthetic conditions, and reading results over $60$ LoCoMo questions whose gold answers appear verbatim in the retrieved blocks. Figure~\ref{fig:memory_attention_analysis} illustrates attention patterns during memory writing and reading across the full-attention layers of the Qwen3.5-4B backbone:
\begin{itemize}
    \item \textbf{Selective Attention during Writing:} As shown in Figure~\ref{fig:memory_attention_analysis}(a), while shallow layers treat factual and filler spans similarly, middle-to-deep layers assign substantially higher attention to factual spans than to filler context. This suggests that the trailing memory tokens preferentially attend to factual content during encoding.
    \item \textbf{Targeted Evidence Grounding during Reading:} Figure~\ref{fig:memory_attention_analysis}(b) demonstrates a clear three-stage progression when generating answers from on-demand decoded KV caches. Shallow layers primarily attend to the \textit{Question}. Attention to the \textit{Answer} block peaks at layers 20--24, exceeding both the question and the other retrieved blocks combined (\textit{Other}). The final layer shifts attention back toward the \textit{Question}. This pattern suggests that the frozen backbone can selectively attend to answer-relevant memory through the decoded KV states.
\end{itemize}

\section{Theoretical Analysis of Memory State Learning}
\label{app:theory}

We provide an information-theoretic and optimization analysis of Eq.~\eqref{eq:memory_state_learning} to explain why existing architectures struggle with long-horizon recall and how MemOnce resolves this bottleneck.

\subsection{Problem Formulation}
Given historical context $x_{1:t-1}$ of total length $N$ and continual input $x_t$, learning an active memory state $M_t = f_\theta(x_t, x_{1:t-1})$ to predict target $y_t$ under a bounded active state budget $C \ll N$ can be formulated as minimizing a variational upper bound on the conditional entropy:
\begin{equation}
    \min_{\theta} \; \underbrace{\mathbb{E}_{(x_{1:t}, y_t)} \!\big[ -\log \pi_\theta(y_t \mid x_t, M_t) \big]}_{\ge\; H(y_t \mid x_t, M_t)} \quad \text{s.t.} \quad M_t = f_\theta(x_t, x_{1:t-1}), \quad \operatorname{Size}(M_t) \le C.
    \label{eq:pib_objective}
\end{equation}
Conditioned on $x_t$, $M_t$ is a deterministic function of $x_{1:t-1}$, forming the Markov chain $y_t \leftrightarrow x_{1:t-1} \to M_t \mid x_t$. By the data processing inequality and the chain rule of mutual information, compressing $x_{1:t-1}$ into $M_t$ induces a non-negative \textit{information loss gap}:
\begin{equation}
    \Delta I(M_t) \triangleq I(x_{1:t-1}; y_t \mid x_t) - I(M_t; y_t \mid x_t) = H(y_t \mid x_t, M_t) - H(y_t \mid x_t, x_{1:t-1}) = I(x_{1:t-1}; y_t \mid x_t, M_t) \ge 0.
    \label{eq:info_gap}
\end{equation}
Minimizing Eq.~\eqref{eq:pib_objective} thus requires both driving $\Delta I(M_t) \to 0$ under $\operatorname{Size}(M_t) \le C$ (so that $M_t$ forms a compact sufficient statistic of $x_{1:t-1}$ with respect to $(x_t, y_t)$) and tightening the variational decoding gap between $\pi_\theta(y_t \mid x_t, M_t)$ and the true posterior.

\begin{table}[!t]
\centering
\caption{\textbf{Comparison of memory state learning paradigms under Eq.~\eqref{eq:pib_objective}.}}
\label{tab:memory_paradigm_comparison}
\resizebox{\textwidth}{!}{
\begin{tabular}{l|cccc}
\toprule
\textbf{Paradigm} & \textbf{Active Memory Mechanism} & \textbf{State Cache} & \textbf{State Optimization} & \textbf{Theoretical Property under Eq.~\eqref{eq:info_gap}} \\
\midrule
Transformer & Full KV cache $\{(K^l, V^l)\}_{l=1}^L$ & $\mathcal{O}(LND)$ & End-to-end $\mathcal{L}_{\text{NTP}}$ & $\Delta I = 0$, but $\mathcal{O}(LND)$ depth-time redundancy \\
Linear / GDN & Recurrent matrix $\{S_N^l \in \mathbb{R}^{d \times d}\}_{l=1}^L$ & $\mathcal{O}(Ld^2)$ & Local $\mathcal{L}_{\text{local}}$ + outer $\mathcal{L}_{\text{NTP}}$ & $\Delta I > 0$ grows with $N$ due to rank bottleneck \\
\textbf{MemOnce (Ours)} & \textbf{Indexed $M_t$ decoded via $u_\theta(M_t)$ on demand} & \textbf{$\mathcal{O}(CD)$} & \textbf{End-to-end $\mathcal{L}_{\text{NTP}}$} & \textbf{$\Delta I \approx 0$ via query-conditioned sparsity} \\
\bottomrule
\end{tabular}
}
\end{table}

\subsection{Theoretical Analysis}

\paragraph{Limitations of Query-Agnostic Memory States.}
As summarized in Table~\ref{tab:memory_paradigm_comparison}, standard architectures construct $M_t = f_\theta(x_{1:t-1})$ independently of $x_t$. Because $M_t$ is fixed prior to observing $x_t$, achieving $\Delta I(M_t) \approx 0$ across all potential future queries requires $M_t$ to preserve the full historical entropy $\Omega(N)$, creating a fundamental dilemma between cache complexity and recall fidelity:
\begin{itemize}
    \item \textit{Cross-Layer Redundancy in Full KV Caches:} Standard Transformers achieve $\Delta I(M_t) = 0$ under next-token prediction ($\mathcal{L}_{\text{NTP}}$) by retaining all intermediate states $M_t^{\text{KV}} = \{(K^l, V^l)\}_{l=1}^L$. However, because residual blocks accumulate contextual representations additively via $H^L = H^0 + \sum_{l=1}^L \operatorname{Block}^l(H^{l-1})$, the final-layer states $H^L$ already consolidate the multi-layer semantic trajectory into a representation directly decodable by lightweight linear projections $u_\theta$, capturing virtually the same usable predictive information ($I(M_t^{\text{KV}}; y_t \mid x_t) \approx I(H^L; y_t \mid x_t)$). Caching all $L$ layers across all $N$ tokens therefore incurs $\mathcal{O}(LND)$ complexity to store redundant intermediate states.
    \item \textit{Rank Bottleneck and Objective Mismatch in Linear Models:} Linear and delta-rule models~\citep{yang2025gated,sun2024learning,lei2026delta} compress $x_{1:t-1}$ into fixed-size matrices $S_N^l \in \mathbb{R}^{d \times d}$ via the online delta rule in Eq.~\eqref{eq:linear_state}. With write gate $\beta_\tau^l$ (acting as the online step size) and erase gate $\alpha_\tau^l$, this recurrence executes one-step online gradient descent on an exponentially weighted local reconstruction loss defined \textit{prior} to observing $x_t$:
    \begin{equation}
        \mathcal{L}_{\text{local}}(S; x_{1:t-1}) \triangleq \sum_{\tau=1}^{N} \beta_\tau^l \left( \prod_{j=\tau+1}^{N} \alpha_j^l \right) \left\| S k_\tau^l - v_\tau^l \right\|_2^2.
        \label{eq:inner_assoc}
    \end{equation}
    Because $S \in \mathbb{R}^{d \times d}$ can losslessly map at most $d$ linearly independent key directions, when $N > d$, the rank constraint $\operatorname{rank}(S_N^l) \le d \ll N$ renders the associative system in Eq.~\eqref{eq:inner_assoc} overdetermined, inducing cross-token interference and exponential decay $\prod_{j=\tau+1}^N \alpha_j^l$ on early states. Crucially, because compression is query-agnostic ($\partial S_N^l / \partial x_t = 0$), task-irrelevant tokens with $I(x_\tau; y_t \mid x_t) = 0$ compete equally for finite rank capacity and overwrite historical signals critical for $y_t$, causing $\Delta I(M_t^{\text{Linear}})$ to grow monotonically with $N$.
\end{itemize}

\paragraph{Query-Conditioned Sufficiency in MemOnce.}
MemOnce overcomes this trade-off through the three-stage factorization $G_t = g_\theta(x_{1:t-1})$, $M_t = i_\theta(x_t, G_t)$, and $C_t = u_\theta(M_t)$ in Eq.~\eqref{eq:three_stages}, decoupling the persistent memory bank $G_t$ from the active memory state $M_t \subseteq G_t$. By the chain rule of mutual information, the total information loss gap in Eq.~\eqref{eq:info_gap} decomposes exactly across our two compression stages:
\begin{equation}
    \Delta I(M_t) = \underbrace{\big[ I(x_{1:t-1}; y_t \mid x_t) - I(G_t; y_t \mid x_t) \big]}_{\Delta I_{\text{write}}(G_t)\text{: compression loss in } g_\theta} + \underbrace{I(G_t \setminus M_t; y_t \mid x_t, M_t)}_{\Delta I_{\text{route}}(M_t)\text{: sparse routing loss in } i_\theta}.
    \label{eq:info_gap_decomp}
\end{equation}
MemOnce drives both terms toward zero while bounding the active cache to $\mathcal{O}(CD)$:
\begin{itemize}
    \item \textit{Depth-Time Compression without Rank Overwriting ($\Delta I_{\text{write}} \approx 0$):} Within each historical segment $x_i$, interleaved linear and causal attention blocks continuously aggregate local sequential context into the trailing final-layer states $h_{i, \,-n_i/r:}^L$. Caching only $G_t \in \mathbb{R}^{\lfloor N/r \rfloor \times D}$ eliminates the layer factor $L$ and compresses sequence length by $r$ without collapsing history into a fixed-rank matrix, empirically preserving $I(G_t; y_t \mid x_t) \approx I(x_{1:t-1}; y_t \mid x_t)$, as verified by our compression ablations (Table~\ref{tab:ablation_compression}).
    \item \textit{Query-Aware Sparse Routing and On-Demand Decoding ($\Delta I_{\text{route}} \approx 0$):} For any realized query $x_t$, predictive information in $G_t$ is highly sparse: only a small subset of historical segments carries non-zero conditional mutual information with $y_t$. By maximizing the InfoNCE contrastive bound in Eq.~\eqref{eq:training_loss} over ground-truth evidence states $m^+ \in G_t^{\text{long}}$, the indexing module $i_\theta$ concentrates task-relevant mutual information into the retrieved subset $M_t \in \mathbb{R}^{C \times D}$ such that unselected states satisfy $I(G_t \setminus M_t; y_t \mid x_t, M_t) \approx 0$. Subsequently, $u_\theta$ decodes $M_t$ on demand into layer-specific KV caches $C_t = \{(K_{\text{Proj}}^l(M_t), V_{\text{Proj}}^l(M_t))\}$ for the attention layers, and end-to-end optimization under $\mathcal{L}_{\text{NTP}}$ directly minimizes the variational upper bound in Eq.~\eqref{eq:pib_objective} without local reconstruction proxies.
\end{itemize}

%% file: main.bbl
\begin{thebibliography}{55}
\providecommand{\natexlab}[1]{#1}
\providecommand{\url}[1]{\texttt{#1}}
\expandafter\ifx\csname urlstyle\endcsname\relax
  \providecommand{\doi}[1]{doi: #1}\else
  \providecommand{\doi}{doi: \begingroup \urlstyle{rm}\Url}\fi

\bibitem[Behrouz et~al.(2025)Behrouz, Zhong, and Mirrokni]{behrouz2025titans}
Ali Behrouz, Peilin Zhong, and Vahab~S. Mirrokni.
\newblock Titans: Learning to memorize at test time.
\newblock \emph{arXiv preprint arXiv:2501.00663}, 2025.

\bibitem[Behrouz et~al.(2026)Behrouz, Li, Deng, Zhong, Razaviyayn, and Mirrokni]{behrouz2026memory}
Ali Behrouz, Zeman Li, Yuan Deng, Peilin Zhong, Meisam Razaviyayn, and Vahab Mirrokni.
\newblock Memory caching: Rnns with growing memory.
\newblock \emph{arXiv preprint arXiv:2602.24281}, 2026.

\bibitem[Beltagy et~al.(2020)Beltagy, Peters, and Cohan]{beltagy2020longformer}
Iz~Beltagy, Matthew~E Peters, and Arman Cohan.
\newblock Longformer: The long-document transformer.
\newblock \emph{arXiv preprint arXiv:2004.05150}, 2020.

\bibitem[Cai et~al.(2026)Cai, Bi, Zhang, Hong, Ge, Huang, Gu, Hu, and Tan]{cai2026last}
Zefan Cai, Sai Bi, Lin Zhang, Yicong Hong, Chongjian Ge, Xun Huang, Jiuxiang Gu, Junjie Hu, and Hao Tan.
\newblock Last layer key-value cache is all you need.
\newblock In \emph{COLM Workshop}, 2026.

\bibitem[Chen et~al.(2024)Chen, Xiao, Zhang, Luo, Lian, and Liu]{chen2024bge}
Jianlv Chen, Shitao Xiao, Peitian Zhang, Kun Luo, Defu Lian, and Zheng Liu.
\newblock Bge m3-embedding: Multi-lingual, multi-functionality, multi-granularity text embeddings through self-knowledge distillation.
\newblock \emph{arXiv:2402.03216}, 2024.

\bibitem[Chhikara et~al.(2025)Chhikara, Khant, Aryan, Singh, and Yadav]{chhikara2025mem0}
Prateek Chhikara, Dev Khant, Saket Aryan, Taranjeet Singh, and Deshraj Yadav.
\newblock Mem0: Building production-ready ai agents with scalable long-term memory.
\newblock \emph{arXiv preprint arXiv:2504.19413}, 2025.

\bibitem[Cobbe et~al.(2021)Cobbe, Kosaraju, Bavarian, Chen, Jun, Kaiser, Plappert, Tworek, Hilton, Nakano, Hesse, and Schulman]{cobbe2021training}
Karl Cobbe, Vineet Kosaraju, Mohammad Bavarian, Mark Chen, Heewoo Jun, Lukasz Kaiser, Matthias Plappert, Jerry Tworek, Jacob Hilton, Reiichiro Nakano, Christopher Hesse, and John Schulman.
\newblock Training verifiers to solve math word problems.
\newblock \emph{arXiv preprint arXiv:2110.14168}, 2021.

\bibitem[Fang et~al.(2026)Fang, Deng, Xu, Jiang, Tang, Xu, Deng, Yao, Wang, Qiao, et~al.]{fang2026lightmem}
Jizhan Fang, Xinle Deng, Haoming Xu, Ziyan Jiang, Yuqi Tang, Ziwen Xu, Shumin Deng, Yunzhi Yao, Mengru Wang, Shuofei Qiao, et~al.
\newblock Lightmem: Lightweight and efficient memory-augmented generation.
\newblock In \emph{ICLR}, 2026.

\bibitem[Gao et~al.(2021)Gao, Tow, Biderman, Black, DiPofi, Foster, Golding, Hsu, McDonell, Muennighoff, et~al.]{gao2021framework}
Leo Gao, Jonathan Tow, Stella Biderman, Sid Black, Anthony DiPofi, Charles Foster, Laurence Golding, Jeffrey Hsu, Kyle McDonell, Niklas Muennighoff, et~al.
\newblock A framework for few-shot language model evaluation.
\newblock \emph{Zenodo}, 2021.

\bibitem[Gao et~al.(2026)Gao, Wei, Zhang, Cheng, Chen, Tang, Jiang, Song, Zhang, Zhao, Yang, Wang, Cao, and Luo]{hysparse2026}
Yizhao Gao, Jianyu Wei, Qihao Zhang, Yu~Cheng, Shimao Chen, Zhengju Tang, Zihan Jiang, Yifan Song, Hailin Zhang, Liang Zhao, Bo~Yang, Gang Wang, Shijie Cao, and Fuli Luo.
\newblock Hysparse: A hybrid sparse attention architecture with oracle token selection and kv cache sharing.
\newblock \emph{arXiv preprint arXiv:2602.03560}, 2026.

\bibitem[Gu and Dao(2023)]{gu2023mamba}
Albert Gu and Tri Dao.
\newblock Mamba: Linear-time sequence modeling with selective state spaces.
\newblock \emph{arXiv preprint arXiv:2312.00752}, 2023.

\bibitem[Hassabis et~al.(2017)Hassabis, Kumaran, Summerfield, and Botvinick]{hassabis2017neuroscience}
Demis Hassabis, Dharshan Kumaran, Christopher Summerfield, and Matthew Botvinick.
\newblock Neuroscience-inspired artificial intelligence.
\newblock \emph{Neuron}, 95\penalty0 (2):\penalty0 245--258, 2017.

\bibitem[Hendrycks et~al.(2021)Hendrycks, Burns, Basart, Zou, Mazeika, Song, and Steinhardt]{hendrycks2021measuring}
Dan Hendrycks, Collin Burns, Steven Basart, Andy Zou, Mantas Mazeika, Dawn Song, and Jacob Steinhardt.
\newblock Measuring massive multitask language understanding.
\newblock In \emph{ICLR}, 2021.

\bibitem[Hu et~al.(2025)Hu, Liu, Yue, Zhang, Liu, Zhu, Lin, Guo, Dou, Xi, et~al.]{hu2025memory}
Yuyang Hu, Shichun Liu, Yanwei Yue, Guibin Zhang, Boyang Liu, Fangyi Zhu, Jiahang Lin, Honglin Guo, Shihan Dou, Zhiheng Xi, et~al.
\newblock Memory in the age of ai agents.
\newblock \emph{arXiv:2512.13564}, 2025.

\bibitem[Jelassi et~al.(2024)Jelassi, Brandfonbrener, Kakade, and Malach]{jelassi2024repeat}
Samy Jelassi, David Brandfonbrener, Sham~M Kakade, and Eran Malach.
\newblock Repeat after me: Transformers are better than state space models at copying.
\newblock \emph{arXiv preprint arXiv:2402.01032}, 2024.

\bibitem[Jiang et~al.(2025)Jiang, Hao, Cho, Li, Yuan, Chen, Ungar, Taylor, and Roth]{jiang2025know}
Bowen Jiang, Zhuoqun Hao, Young-Min Cho, Bryan Li, Yuan Yuan, Sihao Chen, Lyle Ungar, Camillo~J Taylor, and Dan Roth.
\newblock Know me, respond to me: Benchmarking llms for dynamic user profiling and personalized responses at scale.
\newblock In \emph{COLM}, 2025.

\bibitem[{Kimi Team}(2025)]{kimi_team_2025_kimi_linear}
{Kimi Team}.
\newblock Kimi linear: An expressive, efficient attention architecture.
\newblock Technical report, Moonshot AI, 2025.

\bibitem[{Kimi Team} et~al.(2026){Kimi Team}, Bai, Bai, Bao, Cai, Cai, Cao, Cao, Chai, Charles, et~al.]{team2026kimi}
{Kimi Team}, Tongtong Bai, Yifan Bai, Yiping Bao, Jianfeng Cai, Xinyuan Cai, Peizhou Cao, Yuxuan Cao, Ziwei Chai, Y~Charles, et~al.
\newblock Kimi k3: Open frontier intelligence.
\newblock \emph{arXiv preprint arXiv:2607.24653}, 2026.

\bibitem[Lei et~al.(2026)Lei, Zhang, Li, Wang, Fan, Liu, Liu, Ma, Chen, and Poria]{lei2026delta}
Jingdi Lei, Di~Zhang, Junxian Li, Weida Wang, Kaixuan Fan, Xiang Liu, Qihan Liu, Xiaoteng Ma, Baian Chen, and Soujanya Poria.
\newblock {{$\delta$-mem}: Efficient Online Memory for Large Language Models}.
\newblock \emph{arXiv preprint arXiv:2605.12357}, 2026.

\bibitem[Lewis et~al.(2020)Lewis, Perez, Piktus, Petroni, Karpukhin, Goyal, K{\"u}ttler, Lewis, Yih, Rockt{\"a}schel, Riedel, and Kiela]{lewis2020rag}
Patrick Lewis, Ethan Perez, Aleksandra Piktus, Fabio Petroni, Vladimir Karpukhin, Naman Goyal, Heinrich K{\"u}ttler, Mike Lewis, Wen-tau Yih, Tim Rockt{\"a}schel, Sebastian Riedel, and Douwe Kiela.
\newblock Retrieval-augmented generation for knowledge-intensive {NLP} tasks.
\newblock In \emph{NeurIPS}, 2020.

\bibitem[Liu et~al.(2024)Liu, Feng, Wang, Wang, Liu, Zhao, Dengr, Ruan, Dai, Guo, et~al.]{liu2024deepseek}
Aixin Liu, Bei Feng, Bin Wang, Bingxuan Wang, Bo~Liu, Chenggang Zhao, Chengqi Dengr, Chong Ruan, Damai Dai, Daya Guo, et~al.
\newblock Deepseek-v2: A strong, economical, and efficient mixture-of-experts language model.
\newblock \emph{arXiv preprint arXiv:2405.04434}, 2024.

\bibitem[Lozhkov et~al.(2024)Lozhkov, Ben~Allal, von Werra, and Wolf]{lozhkov2024fineweb}
Anton Lozhkov, Loubna Ben~Allal, Leandro von Werra, and Thomas Wolf.
\newblock Fineweb-edu: the finest collection of educational content.
\newblock \emph{Hugging Face Repository}, 2024.

\bibitem[Ma et~al.(2023)Ma, Gong, He, Zhao, and Duan]{ma2023query}
Xinbei Ma, Yeyun Gong, Pengcheng He, Hai Zhao, and Nan Duan.
\newblock Query rewriting in retrieval-augmented large language models.
\newblock In \emph{EMNLP}, 2023.

\bibitem[Maharana et~al.(2024)Maharana, Lee, Tulyakov, Bansal, Barbieri, and Fang]{maharana2024evaluating}
Adyasha Maharana, Dong-Ho Lee, Sergey Tulyakov, Mohit Bansal, Francesco Barbieri, and Yuwei Fang.
\newblock Evaluating very long-term conversational memory of llm agents.
\newblock In \emph{ACL}, 2024.

\bibitem[Merity et~al.(2017)Merity, Xiong, Bradbury, and Socher]{merity2017pointer}
Stephen Merity, Caiming Xiong, James Bradbury, and Richard Socher.
\newblock Pointer sentinel mixture models.
\newblock In \emph{ICLR}, 2017.

\bibitem[Nie et~al.(2026{\natexlab{a}})Nie, Fu, Feng, and Shan]{nie2026evoembedding}
Chang Nie, Chaoyou Fu, Junlan Feng, and Caifeng Shan.
\newblock {EvoEmbedding}: Evolvable representations for long-context retrieval and agentic memory.
\newblock \emph{arXiv preprint arXiv:2606.21649}, 2026{\natexlab{a}}.

\bibitem[Nie et~al.(2026{\natexlab{b}})Nie, Fu, Zhang, Yang, and Shan]{nie2026personavlm}
Chang Nie, Chaoyou Fu, Yifan Zhang, Haihua Yang, and Caifeng Shan.
\newblock Personavlm: Long-term personalized multimodal llms.
\newblock In \emph{CVPR}, 2026{\natexlab{b}}.

\bibitem[Nie et~al.(2026{\natexlab{c}})Nie, Wei, Feng, Fu, and Shan]{nie2026lightomni}
Chang Nie, Jiaju Wei, Junlan Feng, Chaoyou Fu, and Caifeng Shan.
\newblock Light-omni: Reflex over reasoning in agentic video understanding with long-term memory.
\newblock \emph{IEEE TMM}, 2026{\natexlab{c}}.

\bibitem[Paperno et~al.(2016)Paperno, Kruszewski, Lazaridou, Pham, Bernardi, Pezzelle, Baroni, Boleda, and Fern{\'a}ndez]{paperno2016lambada}
Denis Paperno, Germ{\'a}n Kruszewski, Angeliki Lazaridou, Quan~Ngoc Pham, Raffaella Bernardi, Sandro Pezzelle, Marco Baroni, Gemma Boleda, and Raquel Fern{\'a}ndez.
\newblock The {LAMBADA} dataset: Word prediction requiring a broad discourse context.
\newblock In \emph{ACL}, 2016.

\bibitem[Peng et~al.(2023)Peng, Alcaide, Anthony, Albalak, Arcadinho, Biderman, Cao, Cheng, Chung, Derczynski, et~al.]{peng2023rwkv}
Bo~Peng, Eric Alcaide, Quentin Anthony, Alon Albalak, Samuel Arcadinho, Stella Biderman, Huanqi Cao, Xin Cheng, Michael Chung, Leon Derczynski, et~al.
\newblock Rwkv: Reinventing rnns for the transformer era.
\newblock In \emph{EMNLP}, 2023.

\bibitem[Qiao et~al.(2026)]{qiao2026mimov26}
Zongming Qiao et~al.
\newblock Mimo-v2.6: Scaling reinforcement learning towards self-improvement.
\newblock \emph{arXiv preprint arXiv:2610.11959}, 2026.

\bibitem[Qiu et~al.(2026)Qiu, Wang, Li, Li, Xu, Wang, Zhang, Men, Mao, Zhang, et~al.]{qiu2026design}
Zihan Qiu, Zekun Wang, Xiao Li, Yanpeng Li, Yang Xu, Yixuan Wang, Huaqing Zhang, Rui Men, Bochao Mao, Chengruidong Zhang, et~al.
\newblock On the design of qwen3.8-next architecture: Evaluation, efficiency, and training stability.
\newblock \emph{arXiv preprint arXiv:2608.30320}, 2026.

\bibitem[{Qwen Team}(2026)]{qwen35blog}
{Qwen Team}.
\newblock Qwen3.5: Accelerating productivity with native multimodal agents, 2026.
\newblock \url{https://qwen.ai/blog?id=qwen3.5}.

\bibitem[Rein et~al.(2024)Rein, Hou, Stickland, Petty, Pang, Dirani, Michael, and Bowman]{rein2024gpqa}
David Rein, Betty~Li Hou, Asa~Cooper Stickland, Jackson Petty, Richard~Yuanzhe Pang, Julien Dirani, Julian Michael, and Samuel~R. Bowman.
\newblock {GPQA}: A graduate-level google-proof q\&a benchmark.
\newblock In \emph{COLM}, 2024.

\bibitem[Sumers et~al.(2023)Sumers, Yao, Narasimhan, and Griffiths]{sumers2023cognitive}
Theodore~R Sumers, Shunyu Yao, Karthik Narasimhan, and Thomas~L Griffiths.
\newblock Cognitive architectures for language agents.
\newblock \emph{arXiv preprint arXiv:2309.02427}, 2023.

\bibitem[Sun et~al.(2024{\natexlab{a}})Sun, Li, Dalal, Xu, Vikram, Zhang, Dubois, Chen, Wang, Koyejo, et~al.]{sun2024learning}
Yu~Sun, Xinhao Li, Karan Dalal, Jiarui Xu, Arjun Vikram, Genghan Zhang, Yann Dubois, Xinlei Chen, Xiaolong Wang, Sanmi Koyejo, et~al.
\newblock Learning to (learn at test time): {RNNs} with expressive hidden states.
\newblock \emph{arXiv preprint arXiv:2407.04620}, 2024{\natexlab{a}}.

\bibitem[Sun et~al.(2024{\natexlab{b}})Sun, Dong, Zhu, Huang, Wang, Ma, Zhang, Wang, and Wei]{sun2024you}
Yutao Sun, Li~Dong, Yi~Zhu, Shaohan Huang, Wenhui Wang, Shuming Ma, Quanlu Zhang, Jianyong Wang, and Furu Wei.
\newblock You only cache once: Decoder-decoder architectures for language models.
\newblock In \emph{NeurIPS}, 2024{\natexlab{b}}.

\bibitem[Tan et~al.(2025)]{tan2025membench}
Hongda Tan et~al.
\newblock Membench: Towards more comprehensive evaluation on the memory of llm-based agents.
\newblock In \emph{Findings of ACL}, 2025.

\bibitem[Touvron et~al.(2023)]{touvron2023llama}
Hugo Touvron et~al.
\newblock Llama: Open and efficient foundation language models.
\newblock \emph{arXiv preprint arXiv:2302.13971}, 2023.

\bibitem[Vaswani et~al.(2017)Vaswani, Shazeer, Parmar, Uszkoreit, Jones, Gomez, Kaiser, and Polosukhin]{vaswani2017attention}
Ashish Vaswani, Noam Shazeer, Niki Parmar, Jakob Uszkoreit, Llion Jones, Aidan~N Gomez, {\L}ukasz Kaiser, and Illia Polosukhin.
\newblock Attention is all you need.
\newblock In \emph{NeurIPS}, 2017.

\bibitem[Wang et~al.(2024)Wang, Gao, Chen, Jiang, Li, Yang, Yin, Li, Li, Crosby, et~al.]{wang2024memoryllm}
Yu~Wang, Yifan Gao, Xiusi Chen, Haoming Jiang, Shiyang Li, Jingfeng Yang, Qingyu Yin, Zheng Li, Xian Li, Bing Crosby, et~al.
\newblock {MemoryLLM}: Towards self-updatable large language models.
\newblock \emph{arXiv preprint arXiv:2402.04624}, 2024.

\bibitem[Wang et~al.(2025{\natexlab{a}})Wang, Krotov, Hu, Gao, Zhou, McAuley, Gutfreund, Feris, and He]{wang2025m}
Yu~Wang, Dmitry Krotov, Yuanzhe Hu, Yifan Gao, Wangchunshu Zhou, Julian McAuley, Dan Gutfreund, Rogerio Feris, and Zexue He.
\newblock M+: Extending memoryllm with scalable long-term memory.
\newblock In \emph{ICML}, 2025{\natexlab{a}}.

\bibitem[Wang et~al.(2025{\natexlab{b}})Wang, Takanobu, Liang, Mao, Hu, McAuley, and Wu]{wang2025memalpha}
Yu~Wang, Ryuichi Takanobu, Zhiqi Liang, Yuzhen Mao, Yuanzhe Hu, Julian McAuley, and Xiaojian Wu.
\newblock Mem-$\alpha$: Learning memory construction via reinforcement learning.
\newblock \emph{arXiv preprint arXiv:2509.25911}, 2025{\natexlab{b}}.

\bibitem[Wu et~al.(2025)Wu, Wang, Yu, Zhang, Chang, and Yu]{wu2025longmemeval}
Di~Wu, Hongwei Wang, Wenhao Yu, Yuwei Zhang, Kai-Wei Chang, and Dong Yu.
\newblock Longmemeval: Benchmarking chat assistants on long-term interactive memory.
\newblock In \emph{ICLR}, 2025.

\bibitem[Xi et~al.(2025)Xi, Chen, Guo, He, Ding, Hong, Zhang, Wang, Jin, Zhou, et~al.]{xi2025rise}
Zhiheng Xi, Wenxiang Chen, Xin Guo, Wei He, Yiwen Ding, Boyang Hong, Ming Zhang, Junzhe Wang, Senjie Jin, Enyu Zhou, et~al.
\newblock The rise and potential of large language model based agents: A survey.
\newblock \emph{Science China information sciences}, 2025.

\bibitem[Xiaomi(2026)]{xiaomi2026mimo}
LLM-Core Xiaomi.
\newblock Mimo-v2-flash technical report.
\newblock 2026.

\bibitem[Xu et~al.(2026{\natexlab{a}})Xu, Li, Lin, Xue, Xian, Xu, Wu, Zhang, Deng, Yu, et~al.]{xu2026deepseek}
Anyi Xu, B~Li, Bangcai Lin, Bing Xue, BingCheng Xian, Bingzheng Xu, Bochao Wu, Bowei Zhang, Boyi Deng, CC~Yu, et~al.
\newblock Deepseek-v4.1-flash: Pushing the limits of kv cache compression.
\newblock \emph{arXiv preprint arXiv:2609.19969}, 2026{\natexlab{a}}.

\bibitem[Xu et~al.(2026{\natexlab{b}})Xu, Liang, Mei, Gao, Tan, and Zhang]{xu2026mem}
Wujiang Xu, Zujie Liang, Kai Mei, Hang Gao, Juntao Tan, and Yongfeng Zhang.
\newblock A-mem: Agentic memory for llm agents.
\newblock In \emph{NeurIPS}, 2026{\natexlab{b}}.

\bibitem[Yang et~al.(2024)Yang, Wang, Zhang, Shen, and Kim]{yang2024parallelizing}
Songlin Yang, Bailin Wang, Yu~Zhang, Yikang Shen, and Yoon Kim.
\newblock Parallelizing linear transformers with the delta rule over sequence length.
\newblock 2024.

\bibitem[Yang et~al.(2025)Yang, Kautz, and Hatamizadeh]{yang2025gated}
Songlin Yang, Jan Kautz, and Ali Hatamizadeh.
\newblock Gated delta networks: Improving mamba2 with delta rule.
\newblock In \emph{ICLR}, 2025.

\bibitem[Yang et~al.(2018)Yang, Qi, Zhang, Bengio, Salakhutdinov, and Manning]{yang2018hotpotqa}
Zhilin Yang, Peng Qi, Saizheng Zhang, William~W. Bengio, Yoshua~Cohen, Ruslan Salakhutdinov, and Christopher~D. Manning.
\newblock {HotpotQA}: A dataset for diverse, explainable multi-hop question answering.
\newblock In \emph{EMNLP}, 2018.

\bibitem[Yuan et~al.(2025)Yuan, Gao, Dai, Luo, Zhao, Zhang, Xie, Wei, Wang, Ruan, et~al.]{yuan2025native}
Jingyang Yuan, Huazuo Gao, Damai Dai, Junyu Luo, Liang Zhao, Zhengyan Zhang, Zhenda Xie, Yuxing Wei, Lean Wang, Chong Ruan, et~al.
\newblock Native sparse attention: Hardware-aligned and natively trainable sparse attention.
\newblock \emph{arXiv preprint arXiv:2502.11089}, 2025.

\bibitem[Zhang et~al.(2025)Zhang, Li, Long, Zhang, Lin, Yang, Xie, Yang, Liu, Lin, et~al.]{zhang2025qwen3}
Yanzhao Zhang, Mingxin Li, Dingkun Long, Xin Zhang, Huan Lin, Baosong Yang, Pengjun Xie, An~Yang, Dayiheng Liu, Junyang Lin, et~al.
\newblock Qwen3 embedding: Advancing text embedding and reranking through foundation models.
\newblock \emph{arXiv preprint arXiv:2506.05176}, 2025.

\bibitem[Zhang et~al.(2026)Zhang, Guo, Sun, Zhang, Hao, Lin, Zhang, Zhao, Shen, Tang, et~al.]{zhang2026metis}
Zeyu Zhang, Ziliang Guo, Yihang Sun, Xichong Zhang, Xixuan Hao, Zehao Lin, Yang Zhang, Xiaoyan Zhao, Tong Shen, Bo~Tang, et~al.
\newblock Metis: Memory foundation model.
\newblock \emph{arXiv preprint arXiv:2607.26760}, 2026.

\bibitem[Zheng et~al.(2026)Zheng, Shi, Cai, Li, Zhang, Li, Yu, and Ma]{zheng2026lifelong}
Junhao Zheng, Chengming Shi, Xidi Cai, Qiuke Li, Duzhen Zhang, Chenxing Li, Dong Yu, and Qianli Ma.
\newblock Lifelong learning of large language model based agents: A roadmap.
\newblock \emph{IEEE TPAMI}, 2026.

\end{thebibliography}
